\pdfoutput=1

\documentclass[11pt]{article}

\usepackage[]{acl}

\usepackage{times}
\usepackage{latexsym}

\usepackage[T1]{fontenc}

\usepackage[utf8]{inputenc}

\usepackage{microtype}

\usepackage{amsmath,amsfonts,bm}

\def\eqref#1{equation~\ref{#1}}

\def\1{\bm{1}}

\DeclareMathAlphabet{\mathsfit}{\encodingdefault}{\sfdefault}{m}{sl}
\SetMathAlphabet{\mathsfit}{bold}{\encodingdefault}{\sfdefault}{bx}{n}

\usepackage{booktabs}
\usepackage{hyperref}
\usepackage{url}
\usepackage{graphicx}
\graphicspath{{figs/}}
\usepackage{tikz}
\usepackage{subfigure} 
\usepackage{wrapfig}
\usepackage{algorithm}
\usepackage{listings}
\usepackage{pgfplots}
\usetikzlibrary{spy}
\usepackage{color}
\usepackage{xcolor}
\usepackage{comment}
\pgfplotsset{compat=1.17}
\usepackage{multirow}

\definecolor{pink1}{HTML}{F0988C}
\definecolor{pink2}{HTML}{F6CAE5}
\definecolor{red1}{HTML}{f47721}
\definecolor{yellow1}{HTML}{FBCE02}
\definecolor{green1}{HTML}{83D350}
\definecolor{color1}{HTML}{d20962}
\definecolor{color2}{HTML}{f47721}
\definecolor{color3}{HTML}{efdf00}
\definecolor{color4}{HTML}{00a78e}
\definecolor{color5}{HTML}{7ac143}
\definecolor{color6}{HTML}{00bce4}
\definecolor{green2}{HTML}{A9D18E}
\definecolor{blue1}{HTML}{9DC3E6}
\definecolor{green3}{HTML}{00A472}
\definecolor{poscolor1}{HTML}{009ca6}
\definecolor{poscolor2}{HTML}{0689d8}
\definecolor{poscolor3}{HTML}{1428a0}

\title{Adapting Offline Speech Translation Models for Streaming\\with Future-Aware Distillation and Inference}

\author{Biao Fu$^{1,3}$\thanks{\,\, Equal contribution. Work was done during Biao Fu’s research internship at DAMO Academy, Alibaba Group.} , 
Minpeng Liao$^{2}$\footnotemark[1] , 
Kai Fan$^{2}$\footnotemark[1]~~\thanks{\,\, Corresponding author.} ,  
Zhongqiang Huang$^{2}$, \\ 
\textbf{Boxing Chen}$^{2}$, 
\textbf{Yidong Chen}$^{1,3}$~\footnotemark[2] ,
\textbf{Xiaodong Shi}$^{1,3}$ \\
$^{1}$School of Informatics, Xiamen University\\
$^{2}$Alibaba DAMO Academy \\
$^{3}$Key Laboratory of Digital Protection and Intelligent Processing of Intangible Cultural \\ Heritage of Fujian and Taiwan (Xiamen University), Ministry of Culture and Tourism \\
\texttt{biaofu@stu.xmu.edu.cn,\{ydchen,mandel\}@xmu.edu.cn} \\ 
\texttt{\{minpeng.lmp,k.fan,z.huang,boxing.cbx\}@alibaba-inc.com}
}

\begin{document}
\maketitle
\begin{abstract}
A popular approach to streaming speech translation is to employ a single offline model with a \textit{wait-$k$} policy to support different latency requirements, which is simpler than training multiple online models with different latency constraints. However, there is a mismatch problem in using a model trained with complete utterances for streaming inference with partial input. We demonstrate that speech representations extracted at the end of a streaming input are significantly different from those extracted from a complete utterance. To address this issue, we propose a new approach called Future-Aware Streaming Translation (FAST) that adapts an offline ST model for streaming input. FAST includes a Future-Aware Inference (FAI) strategy that incorporates future context through a trainable masked embedding, and a Future-Aware Distillation (FAD) framework that transfers future context from an approximation of full speech to streaming input.
Our experiments on the MuST-C EnDe, EnEs, and EnFr benchmarks show that FAST achieves better trade-offs between translation quality and latency than strong baselines. Extensive analyses suggest that our methods effectively alleviate the aforementioned mismatch problem between offline training and online inference.\footnote{The code is available at \url{https://github.com/biaofuxmu/FAST}}
\end{abstract}
\section{Introduction}

Streaming speech translation (ST) systems generate real-time translations by incrementally processing audio frames, unlike their offline counterparts that have access to complete utterances before translating. 
Typically, streaming ST models use uni-directional encoders \citep{zhang2019lattice,ren-etal-2020-simulspeech,ma-etal-2020-simulmt,zeng-etal-2021-realtrans,zhang2023simple} and are trained with a read/write policy that determines whether to wait for more speech frames or emit target tokens. 
However, it can be expensive to maintain multiple models to satisfy different latency requirements~\citep{zhang-feng-2021-universal,liu2021cross} in real-world applications. 
Recently, some works \citep{papi-etal-2022-simultaneous,dong-etal-2022-learning} have shown that a single offline model with bidirectional encoders (such as Wav2Vec2.0 \cite{NEURIPS2020_92d1e1eb}) can be adapted to streaming scenarios with a \textit{wait-$k$} policy \citep{ma-etal-2019-stacl} to meet different latency requirements and achieve comparable or better performance. 
However, there is an inherent mismatch in using a model bidirectionally trained with complete utterances on incomplete streaming speech during online inference. 

\begin{figure}[!tbp]
\centering
\subfigure[Training: Full speech encoding]{
\label{fig:full}
\includegraphics[width=0.96\linewidth]{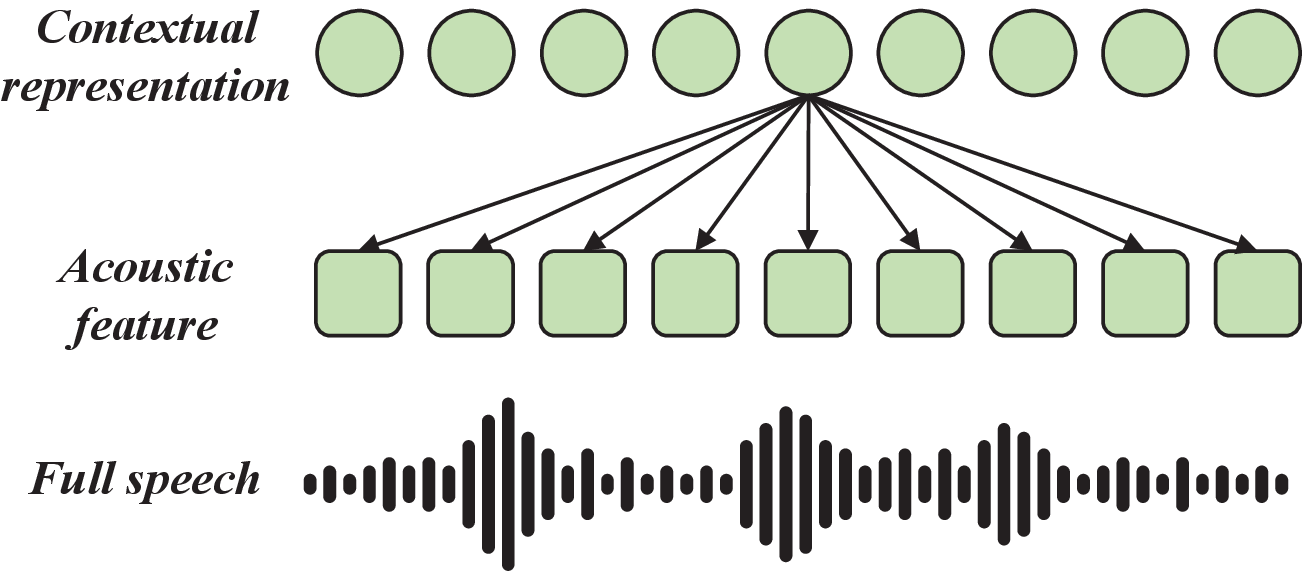}
}
~
\subfigure[Testing: Streaming encoding]{
\label{fig:streaming}
\includegraphics[width=0.96\linewidth]{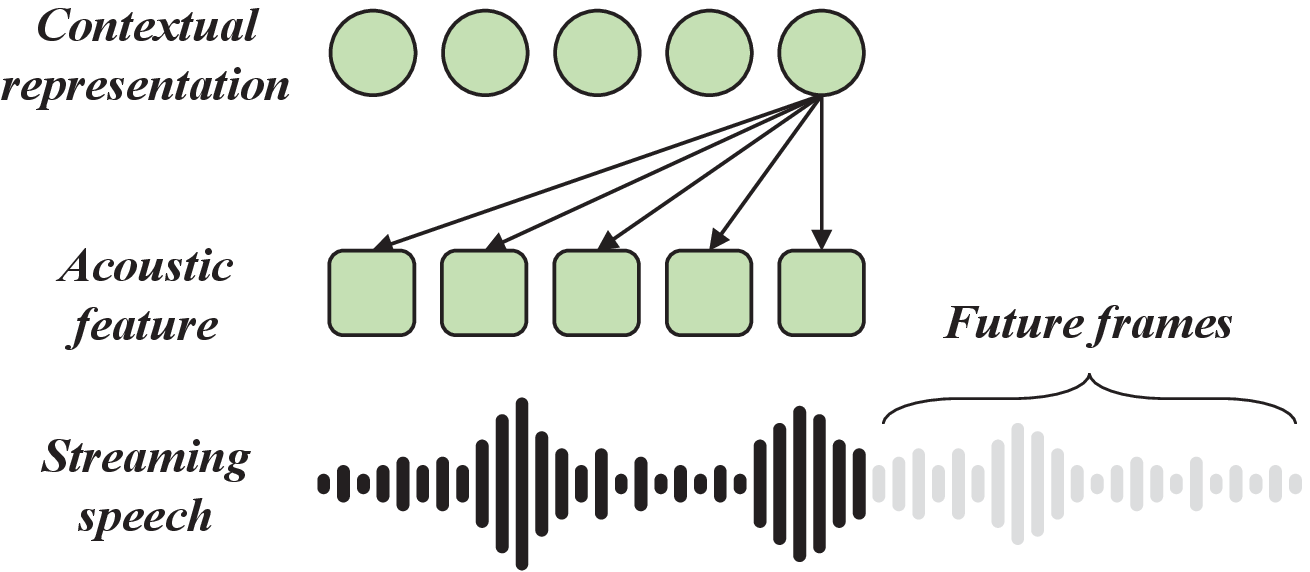}
}
\caption{The input mismatch between offline training and streaming testing.}
\label{fig:mismatch}
\end{figure}

Intuitively, speech representations extracted from streaming inputs (Figure~\ref{fig:streaming}) are less informative than those from full speech encoding (Figure~\ref{fig:full}) due to limited future context, especially toward the end of the streaming inputs, which can be exacerbated by the aforementioned mismatch problem. This raises a natural question: how much do the speech representations differ between the two inference modes? We analyze the gap in speech representations, measured by cosine similarity, at different positions in the streaming input compared to using the full speech (Section \ref{sec:analysis}). We observe a significantly greater gap for representations closer to the end of a streaming segment, with an average similarity score as low as 0.2 for the last frame, and the gap quickly narrows for earlier frames (Figure~\ref{fig:analysis_on_lastpos}). Additionally, we observe more degradation in translation quality for utterances with the greatest gap in speech representations between online and offline inference (see Appendix~\ref{apd:degree}).

We conjecture that the lack of future contexts at the end of streaming inputs can be detrimental to streaming speech translation when using an offline model. 
To this end, we propose a novel Future-Aware Inference (FAI) strategy. 
This approach is inspired by masked language models' ability~\cite{NEURIPS2020_92d1e1eb} to construct representations for masked tokens from their context. 
Specifically, we append a few mask embeddings to the end of the streaming input and leverage the acoustic encoder (Wav2Vec2.0)'s ability to implicitly construct representations for future contexts, which can lead to more accurate representations for the other frames in the streaming input.

Furthermore, we propose a Future-Aware Distillation (FAD) framework that adapts the offline model to extract representations from streaming inputs that more closely resemble those from full speech encoding. 
We expand the original streaming input with two types of future contexts: one with $m$ oracle speech tokens for the teacher model, and another with $m$ mask tokens for the student model, which is initialized from the teacher model. 
We minimize several distillation losses between the output of the teacher and student models. 
By incorporating additional oracle future contexts, the speech representations for the frames in the original streaming input extracted by the teacher model resemble those when the full speech is available. 
FAD aims to adjust the offline model to extract similar representations for streaming input as it would for full speech. 
In combination with FAI, we improve the model's ability to extract quality representations during both training and inference, alleviating the aforementioned mismatch problem. 
We refer to our approach as FAST, which stands for Future-Aware Streaming Translation.

We conducted experiments on the MuST-C EnDe, EnEs, and EnFr benchmarks. 
The results show that our methods outperform several strong baselines in terms of the trade-off between translation quality and latency. 
Particularly, in the lower latency range (when AL is less than 1000\emph{ms}), our approach achieved BLEU improvements of 12 in EnDE, 16 in EnEs, and 14 in EnFr over baseline. 
Extensive analyses demonstrate that our future-aware approach significantly reduces the representation gap between partial streaming encoding and full speech encoding.

\section{Background and Related Work}
 
Speech translation systems can be roughly categorized into non-streaming (offline) and streaming (online) depending on the
inference mode. 
Regardless of the inference mode, speech translation models typically employ the encoder-decoder architecture and are trained on an ST corpus
$\mathcal{D}=\{(\mathbf{x}, \mathbf{z}, \mathbf{y})\}$, where
$\mathbf{x}=(x_1,\ldots, x_{T})$ denotes an audio sequence,
$\mathbf{z}=(z_1,\ldots, z_{I})$ and $\mathbf{y}=(y_1,\ldots, y_{J})$
the corresponding source transcription and target translation
respectively.

\textbf{Non-Streaming Speech Translation} For the non-streaming ST task, the encoder maps the entire input audio $\mathbf{x}$ to the speech representations $\mathbf{h}$, and the decoder generates the $j$-th target token $y_j$ conditional on the full representations $\mathbf{h}$ and the previously generated tokens $y_{<j}$. 
The decoding process of non-streaming ST is defined as $p(\mathbf{y} \mid \mathbf{x})=\prod_{j=1}^{J} p\left(y_{j} \mid \mathbf{x}, \mathbf{y}_{<j}\right)$.

A significant amount of works have focused on non-streaming ST, including pre-training \citep{wang-etal-2020-curriculum,dong2021consecutive,tang-etal-2022-unified,ao-etal-2022-speecht5}, 
multi-task learning \citep{liu2020synchronous,indurthi2020end,indurthi2021task}, 
data augmentation \citep{pino-etal-2019-harnessing,di-gangi-etal-2019-data,mccarthy2020skinaugment}, 
knowledge distillation \citep{dong2021listen,zhao-etal-2021-mutual,du2022regularizing}, 
and cross-modality representation learning \citep{tang-etal-2021-improving,fang-etal-2022-stemm,ye-etal-2022-cross}.

\textbf{Streaming Speech Translation} A streaming ST model generates the $j$-th target token $y_j$ based on streaming audio prefix $\mathbf{x}_{\leq g(j)}$ and the previous tokens $y_{<j}$ , where $g(j)$ is a monotonic non-decreasing function representing the ending timestamp of the audio prefix that needs to be consumed to generate the $j$-th word. 
The decoding probability is calculated as $p(\mathbf{y} \mid \mathbf{x})=\prod_{j=1}^{J} p\left(y_{j} \mid \mathbf{x}_{\leq g(j)}, \mathbf{y}_{<j}\right)$.

Thus, a streaming ST model requires a policy to determine whether to wait for more source speech or emit new target tokens. 
Recent studies \citep{ma-etal-2020-simulmt,ren-etal-2020-simulspeech,zeng-etal-2021-realtrans,dong-etal-2022-learning} make read/write decisions based on a variant of the \textit{wait-$k$} policy that was initially proposed for streaming text translation, which alternates write and read operations after reading the first $k$ source tokens. 
Because there is no explicit word boundaries in a streaming audio, several works attempt to detect word boundaries in the audio sequence by fixed length \citep{ma-etal-2020-simulmt}, Connectionist Temporal Classification \citep{ren-etal-2020-simulspeech,zeng-etal-2021-realtrans,papi-etal-2022-simultaneous}, ASR outputs \citep{chen-etal-2021-direct}, or continuous-integrate-and fire \citep{dong-etal-2022-learning, chang22f_interspeech}. 
Moreover, some studies \citep{arivazhagan-etal-2019-monotonic,Ma2020Monotonic,zhang-etal-2020-learning-adaptive,schneider-waibel-2020-towards,miao-etal-2021-generative,zhang-feng-2022-gaussian,zhang-feng-2022-modeling,zhang-etal-2022-learning,liu-etal-2021-cross,zhang-feng-2022-information,lin2023leapt,zhao2023adaptive} explore adaptive policies to dynamically decide when to read or write for streaming text and/or streaming speech translation. 
\citet{zhang-feng-2022-reducing} fill future source positions with positional encoding as future information during training for simultaneous machine translation (MT) within the prefix-to-prefix framework. 
In this paper, we focus on a matter less attended to -- how to alleviate the mismatch between offline training and online inference.

\textbf{Knowledge Distillation for Streaming Translation}
Existing studies on streaming text and/or speech translation usually introduce future information by distilling sequence-level knowledge from offline MT \cite{ren-etal-2020-simulspeech,zhang2021future,liu-etal-2021-cross,zhu-etal-2022-aisp,deng2023mono4simt,wang2023better} and online MT \cite{Zaidi2021DecisionAR}.
Moreover, \citet{ren-etal-2020-simulspeech} leverage the knowledge from the multiplication of attention weights of streaming ASR and MT models to supervise the attention of the streaming ST model.
However, our FAD aims to reduce the representation gap between full speech and streaming speech.

\section{Preliminary Analysis}
\label{sec:analysis}

In this section, we examine the mismatch problem in Transformer-based \citep{vaswani2017attention} ST architecture between offline training and online decoding. 
In offline full-sentence ST, the speech representation of each frame is obtained by attending to all frames, including future frames, in the transformer encoder layers. 
Recently, a common approach in speech translation is to stack a pre-trained Wav2Vec2.0 \citep{NEURIPS2020_92d1e1eb} as the acoustic encoder with a semantic MT encoder-decoder, resulting in state-of-the-art performance in the ST task \citep{han-etal-2021-learning,dong-etal-2022-learning,fang-etal-2022-stemm,ye-etal-2022-cross}. This approach leverages the ability of Wav2Vec2.0 pre-training to learn better speech representations.

When applying an offline model to streaming inference, the lack of future frames causes an apparent mismatch problem, which can lead to a deterioration in the extracted speech representations. 
To quantify this effect, we examine three offline ST models trained on the MuST-C EnDe dataset using the Chimera \citep{han-etal-2021-learning}, STEMM \citep{fang-etal-2022-stemm}, and MoSST \citep{dong-etal-2022-learning} architectures, with a trainable acoustic encoder initialized from Wav2Vec2.0. 
We conduct analysis on the tst-COMMON set with a duration between 2s and 10s by removing outliers and noisy data, resulting 1829 examples.

\definecolor{posicolor1}{HTML}{4150d8}
\definecolor{posicolor2}{HTML}{28bf7e}
\definecolor{posicolor3}{HTML}{ed7c2f}

\begin{figure}
\centering
\includegraphics[width=1.0\linewidth]{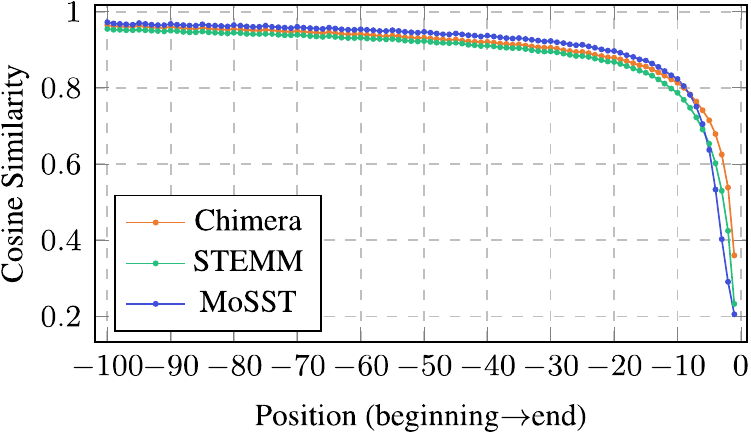}
\caption{The average cosine similarity $\Bar{s}_{-\tau}$ of the end 100 positions in the streaming speech.}
\label{fig:analysis_on_lastpos}
\end{figure}

For an input sequence of audio frames $\mathbf{x}=(x_1,\ldots, x_{T})$, the convolutional subsampler of Wav2Vec2.0 shrinks the length of the raw audio by a factor 320 and outputs the full speech representation sequence $\mathbf{a}$. 
For readability reasons, we uniformly use the notation $T$ to denote the sequence length of $\mathbf{a}=(a_1,\ldots, a_{T})$. 
This simplified notation does not undermine any of our conclusions while making the equations more readable.
For streaming input $\forall t\leq T, \mathbf{\hat{x}}_t=(x_1,\ldots, x_{t})$, Wav2Vec2.0 will output the representation $\mathbf{\hat{a}}_t=(\hat{a}_{t,1},\ldots, \hat{a}_{t,t})$.

To quantify the difference in speech representations between offline and online inputs, we compute the cosine similarity $s_{t,t^{\prime}}$ between the speech representation at the $t^{\prime}$-th ($t^{\prime} \leq t$) position in the streaming audio input $\mathbf{\hat{x}}_t$ and at the same position with full-sentence encoding. 
We then calculate the statistics $\Bar{s}_{-\tau}$ by averaging the cosine similarity over both the testset $\mathcal{B}$ and the time dimension with a reverse index $-\tau$ corresponding to a position $\tau-1$ frames before the end of the streaming input. 
\begin{align}
  &s_{t,t^\prime} (\mathbf{x}) = \operatorname{cos}(\hat{a}_{t,t^{\prime}}, a_{t^{\prime}}), \forall t^\prime\leq t, \label{eq:audio_cos} \\
  &\Bar{s}_{-\tau} = \frac{1}{|\mathcal{B}|}\sum\limits_{\mathbf{x} \in \mathcal{B}} \frac{1}{|\mathbf{x}| - \tau + 1} \sum_{t=\tau}^{|\mathbf{x}|} s_{t,t-\tau+1}(\mathbf{x}) \label{eq:audio_cos2}
\end{align}
Figure \ref{fig:analysis_on_lastpos} displays the $\Bar{s}_{-\tau}$ curve for the last 100 positions in  streaming inputs. 
For $\tau > 10$, the averaged cosine similarity $\Bar{s}_{-\tau}$ is greater than 0.8, indicating that the representations at those positions in a streaming input are similar to those with the full speech. 
However, the curve shows a sharp decline in the averaged cosine similarity $\Bar{s}_{-\tau}$ for the ending positions, particularly for the last one ($\tau=1$), suggesting that the mismatch problem can significantly affect the quality of speech representation for these positions. 
We provide additional analysis in Appendix \ref{apd:additional_analysis}.

\section{Method}

\begin{figure*}
\centering
\includegraphics[width=1.0\linewidth]{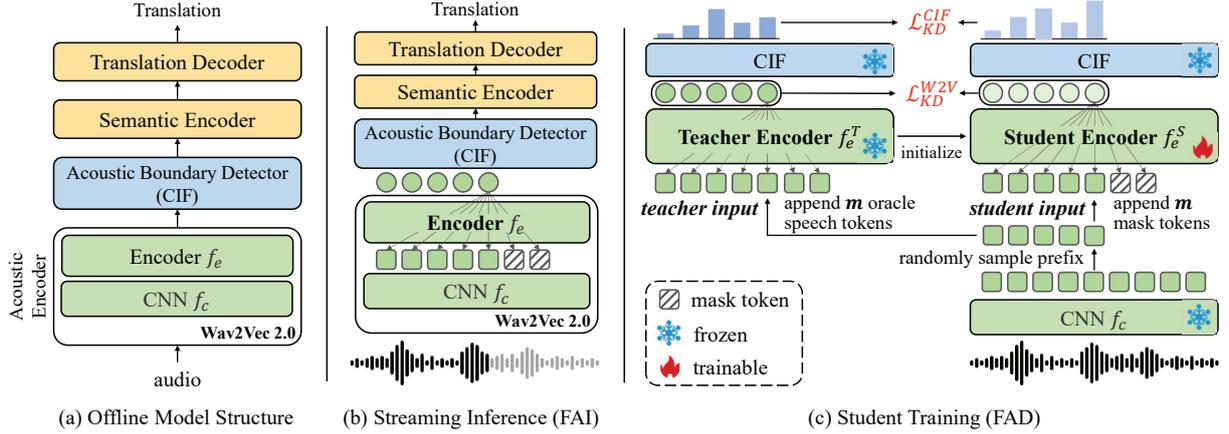}
\caption{Illustration of offline ST model and proposed methods FAI and FAD.}
\label{fig:method}
\end{figure*}

To address the mismatch problem between offline training and online inference, we propose a novel methodology called Future-Aware Streaming Translation (FAST). 
This approach adapts an offline ST model for streaming scenarios by using a Future-Aware Inference (FAI) strategy during inference and a Future-Aware Distillation (FAD) strategy during training. 
An overview of our proposed method is depicted in Figure \ref{fig:method}.

\subsection{Model Architecture}
Unlike previous works \cite{ren-etal-2020-simulspeech,ma-etal-2020-simulmt,zeng-etal-2021-realtrans,liu2021cross} that require training multiple streaming models for different latency requirements, our goal is to train one single offline model to meet the requirements. 
The overall architecture depicted in Figure \ref{fig:method}(a) consists of an acoustic encoder, an acoustic boundary detector, a semantic encoder, and a translation decoder. 
\\
\textbf{Acoustic encoder}: The pre-trained Wav2Vec2.0 is adopted as the acoustic encoder to learn a better speech representation \cite{ye21_interspeech,ye-etal-2022-cross}.
\\
\textbf{Acoustic boundary detector}: To enable the offline ST model to perform chunk-wise streaming inference, we use a Continuous Integrate-and-Fire (CIF) module \cite{dong2022cif} as the acoustic boundary detector to dynamically locate the acoustic boundaries of speech segments following \cite{yi2021efficiently,dong-etal-2022-learning}. 
The CIF module generates an integration weight $\alpha_t$ for each acoustic representation $a_t$ by Wav2Vec2.0. 
Then, CIF accumulates $\alpha_t$ in a step-by-step way.
When the accumulation reaches a certain threshold (e.g. 1.0), the acoustic representations corresponding to these weights are integrated into a single hidden representation $h_j$ by weighted average, indicating a found token boundary. 
The shrunk representations $\mathbf{h}$ will be fed into the semantic encoder. 
To learn the correct acoustic boundaries, we use the source text length $J$ as the weakly supervised signal.
\begin{equation}
\mathcal{L}_{\text{CIF}}=\left\|J-\sum\nolimits_{t=1}^{T}\alpha_t\right\|_2
\label{eq:cif_loss}
\end{equation}
There are two benefits of using CIF as a boundary detector.
For offline ST model, it can address the length gap between speech and text.
It can also provide the acoustic boundaries to perform read/write policies for streaming inference. 
Similar to the word alignment in NMT \cite{li2022structural,li2023neural}, it can align the source audio and source text token.
\\
\textbf{Semantic encoder and Translation decoder}: The standard transformer \cite{vaswani2017attention} composed of $L_{e}$ encoder layers and $L_{d}$ decoder layers is used. 
The translation loss is defined as:
\begin{equation}
\mathcal{L}_{\text{ST}}(\mathbf{x},\mathbf{y})=-\sum\nolimits_{j=1}^{J}\log p\left(y_j \mid y_{<j}, \mathbf{x}\right)
\label{eq:st_loss}
\end{equation}

\subsection{Future-Aware Inference}

The offline ST model is trained with the following objective function:
\begin{equation}
\mathcal{L}_{\text{offline}}=\mathcal{L}_{\text{ST}} + \lambda \cdot \mathcal{L}_{\text{CIF}}
\label{eq:offline_loss}
\end{equation}
where $\lambda$ is a hyper-parameter to balance two losses. 

Based on the analysis in Section \ref{sec:analysis}, we find that it is only necessary for the offline ST model to be aware of a short future during streaming encoding. 
Thus, we first propose a Future-Aware Inference (FAI) strategy to enhance the representations of streaming speech in Figure \ref{fig:method} (b). 

In this strategy, the streaming inference is directly performed on offline ST model without fine-tuning. 
Particularly, we use the mask tokens of Wave2Vec2.0 as the pseudo future context and append them to the speech tokens generated from the already consumed speech frames. 
Because the mask token embedding is trainable when pre-training Wave2Vec2.0, and the contrastive loss is to identify the quantized latent audio representation of masked regions based on unmasked context, this is intuition that mask tokens can possibly encode future context. 
In addition, the masking strategy during pre-training results in approximately 49\% of all time steps being masked with a mean span length of 300ms, it also guarantees that Wav2vec2.0 is able to extract better speech representations even with the presence of large amount of mask tokens.

Wav2Vec2.0 consists of a multi-layer convolutional subsampler $f_c$ and a Transformer encoder $f_e$. 
During our online inference, for each audio prefix $\mathbf{\hat{x}}_t=(x_1,\ldots, x_{t})$, the $f_c$ first outputs streaming speech tokens $\mathbf{\hat{c}}_t=(c_1,\ldots, c_{\tau})$, where $\mathbf{\hat{c}} \in \mathbb{R}^{\tau \times d}$ and $d$ is the dimension of model and $\tau$ is the sequence length after convolutional subsampling.  
Then, we concatenate the streaming speech tokens $\mathbf{\hat{c}}$ and $m$ mask token embeddings $\mathbf{e} \in \mathbb{R}^{d}$ along the time dimension, resulting in a longer sequence of speech tokens $\in \mathbb{R}^{(\tau + m) \times d}$. 
The new speech tokens are then fed into the Transformer encoder $f_e$, but only the first $\tau$ encoder outputs (i.e., speech features) will be kept for the CIF module because, as discussed in Section \ref{sec:analysis}, the last $m$ speech features are of poor quality and adversely affect translation quality. 
Then, if an acoustic boundary is detected by the CIF module, the decoder will emit new words based on \textit{wait-k} policy, otherwise, the streaming speech is continued to be read. 
The FAI strategy is outlined in Algorithm \ref{algo:mask} in Appendix.

\subsection{Future-Aware Distillation}

Although FAI considers mask tokens as the pseudo future context, it is still preferred to leverage the future oracle speech tokens, which is unavailable during inference. 
Therefore, we take one step further by proposing a fine-tuning method -- Future-Aware Distillation (FAD). 
It aims to distill the knowledge from teachers with oracle future contexts into students with pseudo future contexts.

The \textbf{teacher} model is the offline ST by optimizing Eq.~(\ref{eq:offline_loss}) and is frozen. 
The \textbf{student} model has exactly the same architecture as the teacher and is initialised from the teacher. 
However, the semantic encoder and translation decoder are frozen to retain offline-trained ST performance.
\\
\textbf{Training} 
A naive solution is to distill knowledge from the full speech into every possible streaming speech for each audio. 
However, since the length of speech tokens is typically very large, \emph{e.g.}, 300 on average, it is computationally prohibitive. 
To this end, we propose a simple and efficient implementation via random sampling.

Given a full audio waveform $\mathbf{x}$, $f_c$ outputs the speech tokens $\mathbf{c} \in \mathbb{R}^{T \times d}$.
We randomly sample an integer $t \in [1, T]$ to construct the streaming speech token $\mathbf{c}_{\leq t}$.
Then, we define the teacher input of $f_e$ with oracle future context as following:
\begin{equation}
\mathbf{\hat{c}}^\mathcal{T} = \mathbf{c}_{1:t+m} \in \mathbb{R}^{(t+m) \times d},
\end{equation}
where $m$ is a hyper-parameter to denote the number of future contexts. 
The most straightforward approach is to use the full speech as the teacher input. 
However, due to the bidirectional acoustic encoder, the streaming speech representation of the same position constantly changes when consuming new frames.

To maintain consistency with the inference method FAI, we use the mask tokens as the pseudo future context and append them to the sampled speech tokens to construct the student input. 
\begin{equation}
\mathbf{\hat{c}}^\mathcal{S} = \operatorname{Concat}\{\mathbf{c}_{1:t}; m \times [\mathbf{e}]\} \in \mathbb{R}^{(t+m) \times d},
\end{equation}
where $\mathbf{e} \in \mathbb{R}^{d}$ is the mask embedding. 

We can obtain the streaming speech representations from teacher $f_e^\mathcal{T}$ and student $f_e^\mathcal{S}$. 
Then the first $t$ speech representations are fed into the CIF module to derive the teacher and student weight sequence. 
Concretely, they can be written as follows.
\begin{align}
    \mathbf{\hat{a}}^\mathcal{T}, \mathbf{\hat{a}}^\mathcal{S} &= f_e^T(\mathbf{\hat{c}}^\mathcal{T}), f_e^S(\mathbf{\hat{c}}^\mathcal{S}) \\
    \alpha^\mathcal{T}_{1:t}, \alpha^\mathcal{S}_{1:t} &= \operatorname{CIF}(\mathbf{\hat{a}}^\mathcal{T}_{1:t}), \operatorname{CIF}(\mathbf{\hat{a}}^\mathcal{S}_{1:t}) 
\end{align}
Eventually, two distillation losses are proposed to reduce the speech representation gap.
\begin{align}
    \mathcal{L}_{KD}^{\text{W2V}} &= 1-\operatorname{cosine}(\mathbf{\hat{a}}_{1:t}^\mathcal{S},\mathbf{\hat{a}}_{1:t}^\mathcal{T}) \\
    \mathcal{L}_{KD}^{\text{CIF}} &= \sum\nolimits_{\tau=1}^t \operatorname{KL}(\alpha_\tau^\mathcal{T} \| \alpha^\mathcal{S}_\tau)
\end{align}
The first loss is to directly minimize the streaming speech representations with cosine similarity. 
The second loss is to learn more correct acoustic boundaries for online inference by calculating the KL-divergence between two weight distributions. 
Note that according to previous analysis in Section \ref{sec:analysis}, the representations of the first $t$ speech tokens after $f_e^{\mathcal{T}}$ should have high quality if $m > 10$, so only the first $t$ speech representations are taken into account for loss calculation. 

\textbf{Optimization} The total training objective of the FAD can be written as $\mathcal{L}=\mathcal{L}_{KD}^{\text{W2V}} + \mathcal{L}_{KD}^{\text{CIF}}$. 
The overall training procedure of the proposed method is shown in Figure \ref{fig:method}(c).

\section{Experiments}

\definecolor{maincolor}{HTML}{A955FF}
\definecolor{maincolor2}{HTML}{FFA200}
\definecolor{maincolor3}{HTML}{EE005F}
\definecolor{maincolor4}{HTML}{8E3400}

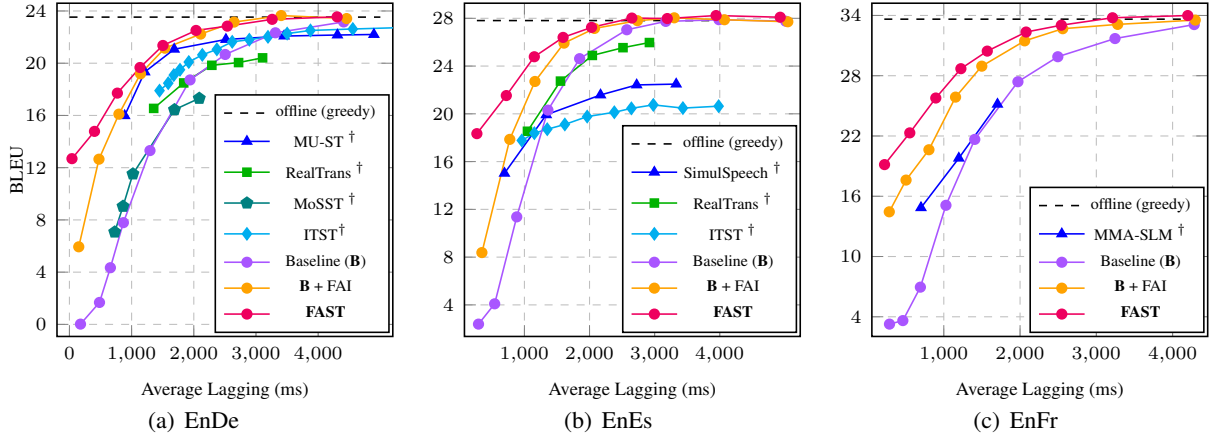
\begin{figure*}[t]
\pgfplotsset{width=6cm,height=6cm,
    every axis y label/.append style={at={(-0.07,0.5)}},
    every axis/.append style={line width=0.6pt},
}
\centering
\subfigure[EnDe]{
\begin{tikzpicture}[baseline]
\begin{axis}[
    ylabel=BLEU,
    xlabel=Average Lagging (ms),
    enlargelimits=0.04,
    font=\scriptsize,
    legend style={font=\tiny,
    at={(0.73,0.01)},
    anchor=south,
    legend columns=1},
    xmajorgrids=true,
    ymajorgrids=true,
    grid style=dashed,
    xmin=0, xmax = 5000,
    xtick={0,1000,2000,3000,4000},
    ytick={0,4,8,12,16,20,24},
]
\addplot[color=black,dashed,line width=0.6pt] coordinates {(1,23.53)(5000,23.53)};
\addplot[color=blue,mark=triangle*,line width=0.6pt] coordinates{(888,15.98)(1211,19.33)(1685,21.07)(2516,21.82)(3452,22.08)(4310,22.16)(4896,22.20)};
\addplot[color=green!70!black,mark=square*, mark size=1.5pt,line width=0.6pt] coordinates {(1355,16.54)(1838,18.49)(2290,19.84)(2720,20.05)(3106,20.41)};
\addplot[color=teal,mark=pentagon*, mark size=2.2pt,line width=0.6pt] coordinates {(728,7.07)(862,9.04)(1021,11.52)(1689,16.44)(2088,17.31)}; 
\addplot[color=cyan,mark=diamond*, mark size=2.2pt,line width=0.6pt] coordinates {(1449,17.9)(1589,18.47)(1678,19.09)(1778,19.5)(1919,20.09)(2137,20.64)(2371,21.06)(2618,21.64)(2893,21.8)(3193,22.02)(3501,22.27)(3876,22.51)(4557,22.62)(5206,22.71)}; 
\addplot[color=maincolor,mark=*, mark size=1.8pt,line width=0.6pt] coordinates {(178,0.02)(483.4,1.68)(658.6,4.34)(867.43,7.79)(1295.16,13.31)(1939.46,18.72)(2505.04,20.67)(3312.15,22.33)(4409.65,23.16)}; 
\addplot[color=maincolor2,mark=*, mark size=1.8pt,line width=0.6pt] coordinates {(150,5.94)(475,12.65)(795.84,16.1)(1143.37,19.19)(1533.63,21.15)(2109.41,22.23)(2647.03,23.15)(3403.9,23.65)(4457.14,23.42)}; 
\addplot[color=maincolor3,mark=*, mark size=1.8pt,line width=0.6pt] coordinates {(41,12.69)(403,14.78)(771,17.71)(1135,19.67)(1503,21.36)(2036,22.51)(2539,22.84)(3260,23.36)(4305,23.55)};

\legend{offline (greedy),MU-ST $^{\dagger}$,RealTrans $^{\dagger}$,MoSST $^{\dagger}$, ITST$^{\dagger}$,Baseline (\textbf{B}),\textbf{B} + FAI,\textbf{FAST}}
\end{axis}
\end{tikzpicture}}
~
\subfigure[EnEs]{
\begin{tikzpicture}[baseline]
\begin{axis}[
    xlabel=Average Lagging (ms),
    enlargelimits=0.04,
    font=\scriptsize,
    legend style={font=\tiny,
    at={(0.73,0.01)},
    anchor=south,
    legend columns=1},
    xmajorgrids=true,
    ymajorgrids=true,
    grid style=dashed,
    xtick={0,1000,2000,3000,4000},
    ytick={0,4,8,12,16,20,24,28},
]
\addplot[color=black,dashed,line width=0.6pt] coordinates {(270,27.81)(4000,27.81)};
\addplot[color=blue,mark=triangle*,line width=0.6pt] coordinates{(694,15.02)(1336,19.92)(2169,21.58)(2724,22.42)(3331,22.49)};
\addplot[color=green!70!black,mark=square*, mark size=1.5pt,line width=0.6pt] coordinates {(1047,18.54)(1554,22.74)(2043,24.89)(2514,25.54)(2920,25.97)};
\addplot[color=cyan,mark=diamond*, mark size=2.2pt,line width=0.6pt] coordinates {(960,17.77)(1153,18.38)(1351,18.71)(1621,19.11)(1964,19.77)(2381,20.13)(2643,20.46)(2980,20.75)(3434,20.48)(3983,20.64)};
\addplot[color=maincolor,mark=*, mark size=1.8pt,line width=0.6pt] coordinates {(295,2.39)(543,4.09)(882,11.37)(1361,20.31)(1848,24.62)(2572,27.04)(3171,27.74)(3988,27.88)(5012,27.76)}; 
\addplot[color=maincolor2,mark=*, mark size=1.8pt,line width=0.6pt] coordinates {(347,8.38)(775,17.86)(1162,22.71)(1608,25.92)(2076,27.15)(2736,27.8)(3301,28.04)(4072,27.88)(5045,27.71)}; 
\addplot[color=maincolor3,mark=*, mark size=1.8pt,line width=0.6pt] coordinates {(270,18.34)(722,21.53)(1152,24.78)(1594,26.4)(2031,27.24)(2650,28.02)(3194,27.98)(3943,28.23)(4928,28.09)}; 

\legend{offline (greedy),SimulSpeech $^{\dagger}$,RealTrans $^{\dagger}$,ITST $^{\dagger}$,Baseline (\textbf{B}),\textbf{B} + FAI,\textbf{FAST}}
\end{axis}
\end{tikzpicture}}
~
\subfigure[EnFr]{
\begin{tikzpicture}[baseline]
\begin{axis}[
    xlabel=Average Lagging (ms),
    enlargelimits=0.04,
    font=\scriptsize,
    legend style={font=\tiny,
    at={(0.73,0.01)},
    anchor=south,
    legend columns=1},
    xmajorgrids=true,
    ymajorgrids=true,
    grid style=dashed,
    ytick={4,10,16,22,28,34},
]
\addplot[color=black,dashed,line width=0.6pt] coordinates {(220,33.63)(4200,33.63)};
\addplot[color=blue,mark=triangle*,line width=0.6pt] coordinates{(701,14.86)(1197,19.79)(1704,25.16)};
\addplot[color=maincolor,mark=*, mark size=1.8pt,line width=0.6pt] coordinates {(288,3.27)(463,3.62)(693,6.95)(1028,15.1)(1406,21.66)(1972,27.4)(2495,29.89)(3245,31.7)(4283,33.09)}; 
\addplot[color=maincolor2,mark=*, mark size=1.8pt,line width=0.6pt] coordinates {(285,14.45)(505,17.61)(805,20.63)(1154,25.87)(1498,28.95)(2060,31.47)(2559,32.68)(3280,33.11)(4297,33.54)}; 
\addplot[color=maincolor3,mark=*, mark size=1.8pt,line width=0.6pt] coordinates {(223,19.15)(554,22.31)(895,25.78)(1224,28.7)(1570,30.45)(2079,32.35)(2541,33.03)(3212,33.77)(4199,33.99)}; 

\legend{offline (greedy),MMA-SLM $^{\dagger}$, Baseline (\textbf{B}),\textbf{B} + FAI,\textbf{FAST}}

\end{axis}
\end{tikzpicture}}
\caption{The translation quality (BLEU) against the latency metrics (AL) on the tst-COMMON set of MuST-C EnDe, EnEs, and EnFr dataset. 
$^{\dagger}$ denotes that the results are obtained from corresponding papers.
offline is the offline performance of teacher model (offline-trained ST) by greedy search. 
The curve corresponding to \textbf{B} is the online performance of the teacher model using vanilla \textit{wait-k} policy. 
The curve corresponding to \textbf{B} + FAI is the online performance of the teacher model with our FAI strategy.
The curve corresponding to \textbf{FAST} is the online performance of our student model with the FAI strategy, \emph{i.e.}, FAD + FAI.}
\label{fig:main_results}
\end{figure*}
\subsection{Experimental Settings}

\textbf{Datasets} We evaluate our approach on MuST-C V1 English-German (EnDe), English-Spanish (EnEs) and English-French (EnFr) datasets \citep{di-gangi-etal-2019-must}, 
where limited previous works discussed the En-Fr streaming ST with BLEU-latency curve. 
All the corpora contain source audios, source transcriptions, and target translations, and the results reported are conducted on the corresponding tst-COMMON set. 
Detailed statistics of different language pairs are given in Appendix \ref{apd:data_statistics}.

For speech data, we normalize the raw audio wave to the range of $[-1,1)$. 
For text data, we keep punctuation and remove non-printing characters, and remain case-sensitive. 
For each translation direction, the unigram SentencePiece\footnote{\url{https://github.com/google/sentencepiece}} model \citep{kudo-richardson-2018-sentencepiece} is used to learn a shared vocabulary of size 10k. 
\\
\textbf{Model Configuration} 
For the acoustic encoder, we use Wav2vec2.0\footnote{\url{https://dl.fbaipublicfiles.com/fairseq/wav2vec/wav2vec\_small.pt}} \citep{NEURIPS2020_92d1e1eb} following the base configurations. 
We construct the acoustic boundary detector by applying the CIF \citep{yi2021efficiently} on the last dimension of speech representation. 
We use 8 and 6 layers for the semantic encoder and the translation decoder respectively, with 4 attention heads and 768 hidden units.
\\
\textbf{Training} 
The detailed training schedule of the offline ST model is given in Appendix \ref{apd:details_of_training}. 
We set the length $m$ of future context tokens to 50 for both FAD and FAI. 
All hyper-parameters are tuned on EnDe devset and applied to other language pairs. 
We train all models with 3.2 million frames per batch on 8 Nvidia Tesla V100 GPUs. 
We implement our models with Fairseq\footnote{\url{https://github.com/pytorch/fairseq}} \cite{ott-etal-2019-fairseq}.
\\
\textbf{Inference} We average the checkpoints of the best 10 epochs on development set for evaluation. 
We perform streaming-testing with the \textit{wait-$k$} policy. 
$k$ is counted by the detected acoustic units from the CIF module. 
To follow the tradition in simultaneous translation \citep{zeng-etal-2021-realtrans,dong-etal-2022-learning}, we do not rewrite the tokens that have already been generated. 
\\
\textbf{Evaluation Metrics} We use SacreBLEU\footnote{\url{https://github.com/mjpost/sacrebleu}} for the translation quality. 
The latency is evaluated with Average Latency (AL) \citep{ma-etal-2019-stacl}, Average Proportion (AP) \citep{cho2016can}, and Differentiable Average Lagging (DAL) \citep{cherry2019thinking} in the SimulEval\footnote{\url{https://github.com/facebookresearch/SimulEval}} \citep{ma-etal-2020-simuleval}. 
\\
\textbf{System Settings} 
We compare our method with several strong end-to-end streaming ST approaches. 
(\romannumeral1) \textit{SimulSpeech} \citep{ren-etal-2020-simulspeech} and \textit{RealTranS} \citep{zeng-etal-2021-realtrans} use uni-directional encoder rather than bidirectional one. 
(\romannumeral2) \textit{MoSST} \citep{dong-etal-2022-learning} applies an offline-trained model with a monotonic segmentation module for streaming testing and achieves competitive performance. 
(\romannumeral3) \textit{MMA-SLM} \cite{indurthi-etal-2022-language} enhances monotonic attention to make better read/write decisions by integrating future information from language models.
(\romannumeral4) \textit{ITST} \cite{zhang-feng-2022-information} learns an adaptive read/write policy by quantifying the transported information weight from source token to the target token.
(\romannumeral5) \textit{MU-ST} \citep{zhang-etal-2022-learning} learns an adaptive segmentation policy to detect meaningful units, which makes read/write decisions. 
(\romannumeral6) \textit{Baseline} is our offline-trained ST model (\textbf{B} for abbreviation). 
For fair comparisons, it has the same structure as MoSST.

\subsection{Main Results}

We presents the main results in Figure \ref{fig:main_results} \footnote{The extended results for other latency metrics (AP and DAL) are described in Appendix \ref{apd:appendix_more_results}.}. 
Compared with the online models SimulSpeech, RealTranS, and ITST, our offline model (baseline) achieves higher translation quality with high latency as it encodes bidirectional context information during training, however, in the low latency region, it performs poorly due to the input mismatch between offline-training and online-decoding. 

\textbf{B + FAI} With the ability to reduce this mismatch, FAI is directly applied for our offline (baseline) model and can achieve higher BLEU in all latency regions. 
In particular, it outperforms our most compatible baseline \textbf{B} by large margins in lower latency regions (when AL is less than 1000\emph{ms}), with improvements over 6 BLEU in both EnDe and EnEs, 10 BLEU in EnFr.

\textbf{FAST} (FAD + FAI) Furthermore, our FAST achieves the best trade-off between translation quality and latency, especially at extremely low latency region (AL is about 200\emph{ms}, $k=1$), achieving the improvements of 6 BLEU in EnDe, 10 BLEU in EnEs, and 4 BLEU in EnFr compared to B + FAI.
It indicates that FAST can effectively mitigate the input mismatch between offline-training and online-decoding.
In addition, our method achieves comparable translation quality with full-speech translation at middle latency (at AL around 3000\emph{ms}), especially for EnEs.

\subsection{Ablation Study}
\label{sec:ablation}

\begin{figure}[t]
\centering
\includegraphics[width=1.0\linewidth]{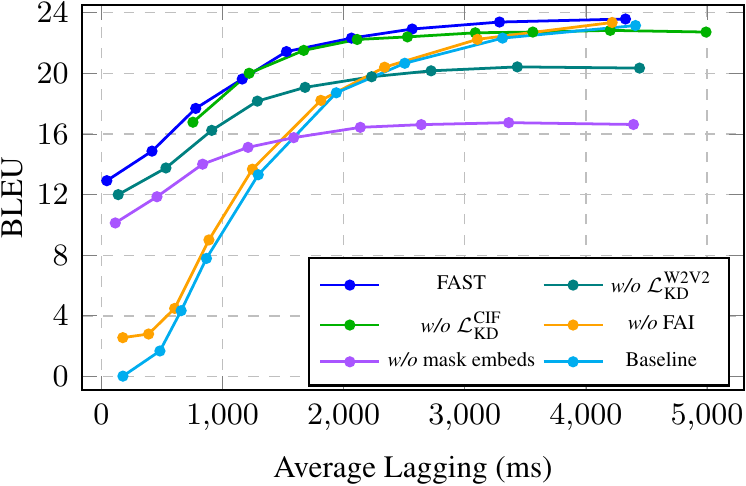}
\caption{Ablation study of our method on the tst-COMMON set of MuST-C EnDe dataset. The observed points in the plots represent wait-$k$ policy with $k=\{1,3,5,7,9,12,15,20,30\}$.}
\label{fig:ablation}
\end{figure}

In this section, we study the effectiveness of our methods. 
All ablation results are obtained from the MuST-C EnDe tst-COMMON set. 
The results are shown in Figure \ref{fig:ablation}.

(1) \textit{w/o $\mathcal{L}_{KD}^{\text{W2V2}}$}: if removing the $\mathcal{L}_{KD}^{\text{W2V2}}$, the translation quality drops by 1-2 BLEU in all latency regions, including high latency region.
This demonstrates optimizing $\mathcal{L}_{KD}^{\text{W2V2}}$ can guarantee the full speech translation performance.

(2) \textit{w/o $\mathcal{L}_{KD}^{\text{CIF}}$}: If removing the $\mathcal{L}_{KD}^{\text{CIF}}$, the translation quality will be slightly degraded. 
However, we observe that the distances between two consecutive acoustic boundaries become larger. 
For example, the AL of this variant at \textit{wait-1} is greater than 750, but the AL of the other variants at \textit{wait-1} is approximately 150. 
As expected, optimizing $\mathcal{L}_{KD}^{\text{CIF}}$ can ensure the correct acoustic boundaries.

(3) \textit{w/o FAI}: In this variant, we use the student model by FAD with vanilla \textit{wait-k} policy for streaming inference (\emph{i.e.}, inference without mask tokens). 
However, FAD training considers mask tokens as student input, so this mismatch leads to significant performance degradation in low and middle latency regions.  
This indicates that our FAD and FAI should be used together to achieve better streaming performance.

(4) \textit{w/o mask embeddings}: During training and inference, our model appends $m$ mask tokens into streaming speech tokens as the pseudo future contexts. 
In this variant, we remove the mask tokens during both training and inference. 
Even though no mismatch, we still observe a significant drop in translation quality, especially for high latency. 
This result indicates that the pseudo future contexts can enhance the streaming speech representations.

\subsection{How much future context is needed?}

\definecolor{cycle1}{HTML}{1f77b4}
\definecolor{cycle2}{HTML}{ff7f0e}
\definecolor{cycle3}{HTML}{2ca02c}
\definecolor{cycle4}{HTML}{d62728}
\definecolor{cycle5}{HTML}{9467bd}
\definecolor{cycle6}{HTML}{8c564b}
\definecolor{cycle7}{HTML}{e377c2}
\definecolor{cycle8}{HTML}{7f7f7f}
\definecolor{cycle9}{HTML}{bcbd22}
\definecolor{cycle10}{HTML}{17becf}

\begin{figure}[t]
\centering
\includegraphics[width=1.0\linewidth]{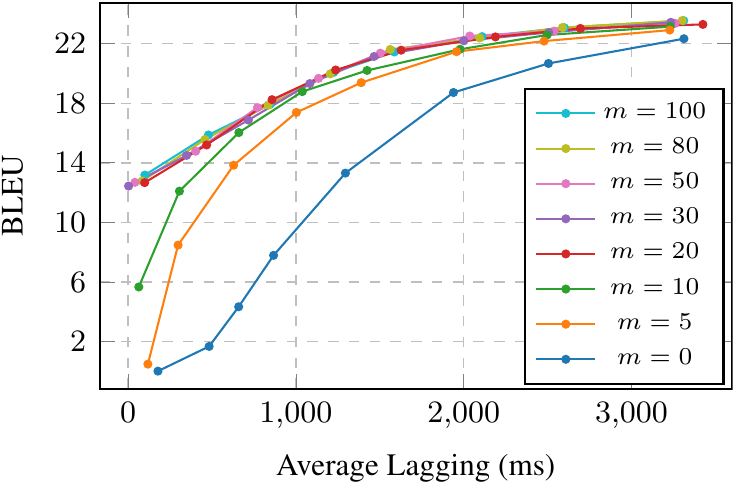}
\caption{Effect on BLEU-AL curve of FAST w.r.t. $m$.} 
\label{fig:length_bleu}
\end{figure}

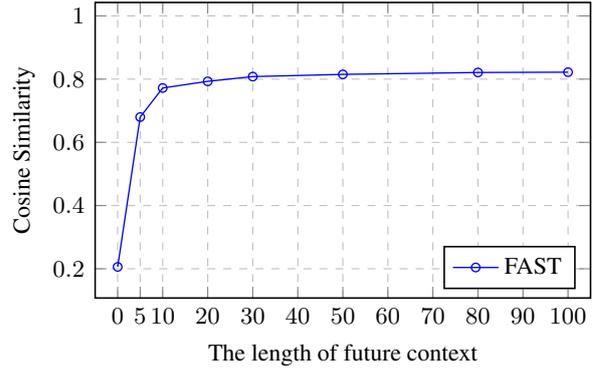
\begin{figure}[t]
\centering
\pgfplotsset{width=8.1cm,height=5.5cm,
    every axis y label/.append style={at={(-0.1,0.5)}},
    every axis/.append style={line width=0.6pt},
}
\begin{tikzpicture}[baseline]
\begin{axis}[
    ylabel=Cosine Similarity,
    xlabel=The length of future context,
    enlargelimits=0.05,
    legend pos=south east,
    legend style={font=\small},
    font=\small,
    ymajorgrids=true,
    xmajorgrids=true,
    grid style=dashed,
    ymin=0.15, ymax = 1.0,
    xtick={0,5,10,20,30,40,50,60,70,80,90,100},
]
\addplot[color=blue,mark=o, mark size=1.6pt,line width=0.5pt] coordinates{(0,0.206)(5,0.68)(10,0.772)(20,0.793)(30,0.808)(50,0.815)(80,0.821)(100,0.822)};
\legend{FAST,FAI}
\end{axis}
\end{tikzpicture}
\caption{Effect on the $\Bar{s}_{-1}$ w.r.t. $m$.}
\label{fig:length_cos}
\end{figure}

To answer this question, we explore the FAST (FAD + FAI) with different lengths of future context. 
Figure \ref{fig:length_bleu} shows the overall results. 
$m=0$ means the offline system without distillation. 
The offline system inherits the mismatch problem, but our method gradually improves the performance as $m$ increasing from 0 to 20. 
Since we found only the representation of last 10 positions is poor (in Section \ref{sec:analysis}), FAST obtains similar BLEU-AL curve when $m$ is significantly larger than 10, \emph{e.g.}, 20-100. 

After the FAD training, we investigate the representation of the last position (before mask tokens) by $\bar{s}_{-1}$ in Eq.~(\ref{eq:audio_cos2}) w.r.t. $m$. 
The results are shown in Figure \ref{fig:length_cos}.
We observe that 1) as $m$ increases, the streaming speech representation of the last position becomes better; 
2) the curves of the cosine similarity becomes flattened when $m > 10$ significantly. 
This is consistent with the trend in Figure \ref{fig:length_bleu}.

\subsection{Analysis on The Representation Gap}

\definecolor{poscolor4}{HTML}{38ada9}
\definecolor{poscolor5}{HTML}{78e08f}
\definecolor{poscolor6}{HTML}{b8e994}
\definecolor{poscolor7}{HTML}{008BCC}
\definecolor{poscolor8}{HTML}{00AC53}
\definecolor{poscolor9}{HTML}{3E4372}
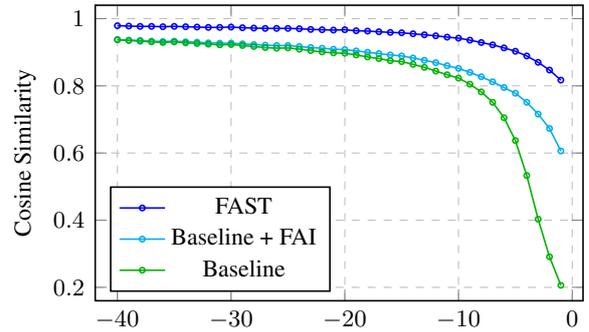
\begin{figure}[t]
\centering
\pgfplotsset{
    width=8cm,height=5.5cm,
    every axis y label/.append style={at={(-0.1,0.5)}},
    every axis/.append style={line width=0.6pt},
}

\begin{tikzpicture}[baseline]
\begin{axis}[
    ylabel=Cosine Similarity,
    enlargelimits=0.05,
    font=\small,
    legend pos=south west,
    legend style={font=\small},
    ymajorgrids=true,
    xmajorgrids=true,
    grid style=dashed,
    ymin=0.2, ymax = 1.0,
    ytick={0.2,0.4,0.6,0.8,1.0},
]
\addplot[color=blue,mark=o, mark size=1pt] coordinates{(-40,0.979)(-39,0.978)(-38,0.977)(-37,0.977)(-36,0.976)(-35,0.977)(-34,0.976)(-33,0.975)(-32,0.974)(-31,0.974)(-30,0.975)(-29,0.973)(-28,0.972)(-27,0.971)(-26,0.971)(-25,0.971)(-24,0.97)(-23,0.968)(-22,0.967)(-21,0.966)(-20,0.967)(-19,0.964)(-18,0.963)(-17,0.961)(-16,0.959)(-15,0.958)(-14,0.955)(-13,0.952)(-12,0.949)(-11,0.945)(-10,0.942)(-9,0.936)(-8,0.929)(-7,0.922)(-6,0.913)(-5,0.903)(-4,0.889)(-3,0.87)(-2,0.847)(-1,0.817)};
\addplot[color=cyan,mark=o, mark size=1pt] coordinates{(-40,0.938)(-39,0.936)(-38,0.935)(-37,0.933)(-36,0.932)(-35,0.933)(-34,0.931)(-33,0.930)(-32,0.928)(-31,0.927)(-30,0.927)(-29,0.925)(-28,0.923)(-27,0.921)(-26,0.920)(-25,0.920)(-24,0.917)(-23,0.914)(-22,0.912)(-21,0.909)(-20,0.908)(-19,0.904)(-18,0.900)(-17,0.896)(-16,0.892)(-15,0.889)(-14,0.883)(-13,0.876)(-12,0.869)(-11,0.860)(-10,0.852)(-9,0.840)(-8,0.827)(-7,0.812)(-6,0.795)(-5,0.778)(-4,0.751)(-3,0.716)(-2,0.673)(-1,0.606)};
\addplot[color=green!70!black,mark=o, mark size=1pt] coordinates{(-40,0.937)(-39,0.935)(-38,0.933)(-37,0.931)(-36,0.929)(-35,0.931)(-34,0.928)(-33,0.926)(-32,0.924)(-31,0.922)(-30,0.923)(-29,0.920)(-28,0.917)(-27,0.914)(-26,0.912)(-25,0.913)(-24,0.909)(-23,0.905)(-22,0.901)(-21,0.898)(-20,0.897)(-19,0.892)(-18,0.886)(-17,0.881)(-16,0.875)(-15,0.872)(-14,0.864)(-13,0.855)(-12,0.844)(-11,0.833)(-10,0.823)(-9,0.805)(-8,0.782)(-7,0.751)(-6,0.705)(-5,0.637)(-4,0.533)(-3,0.403)(-2,0.291)(-1,0.206)};
\legend{FAST,Baseline + FAI,Baseline,}
\end{axis}
\end{tikzpicture}
\caption{Effect on the average cosine similarity $\Bar{s}_{-t^{\prime}}$ of the streaming speech representations at the end positions (before mask tokens). After applying FAI and FAST, the representations of the end positions are improved.}
\label{fig:whywork}
\end{figure}

Figure \ref{fig:whywork} plots the changes of average cosine similarity $\Bar{s}_{-t^{\prime}}$ in Eq.~(\ref{eq:audio_cos2}) of the last 40 positions (before mask tokens) in the streaming speech after applying the FAI or FAST (FAD + FAI). 
They achieve at least 0.6 and 0.8 cosine similarity at the last position, respectively. 
The baseline only has the $<0.6$ cosine similarity for the last 4 positions and only 0.2 for the last position. 
It indicates that the representations with FAI are closer to those of the full speech, especially at the ending positions, and FAD training can further close this gap.

\subsection{What examples are improved?}
\label{sec:monotonic}
\begin{table}[t]
\begin{center}
\resizebox{\linewidth}{!}{
\begin{tabular}{l|l|l|l|c}
\toprule
Method & \multicolumn{1}{c|}{\textbf{Easy}}  & \multicolumn{1}{c|}{\textbf{Medium}} & \multicolumn{1}{c|}{\textbf{Hard}}  & \textbf{AL}   \\ \midrule
Offline (greedy) & 26.38 & 23.22 & 21.26 & -     \\
Baseline            & 18.88 & 12.95 & 10.38  & 1295 \\ %
$+$ FAI        & 23.88$^{+5.00}$ & 18.99$^{+6.04}$ & 16.45$^{+6.07}$  & 1143 \\ 
FAST         & 24.44$^{+5.56}$ & 19.89$^{+6.94}$ & 16.53$^{+6.15}$  & 1135 \\\bottomrule
\end{tabular}}
\end{center}
\caption{Performance (BLEU) on different monotonic levels on test set of MuST-C EnDe.}
\label{tab:reorder}
\end{table}

For tst-COMMON on MuST-C EnDe, we use awesome-align\footnote{\url{https://github.com/neulab/awesome-align}} \cite{dou-neubig-2021-word} to identify the token-level alignment between source transcription and target translation following \citet{zhang-feng-2022-reducing}. 
First, we define the source-to-target alignment position shift as $\max\{0, i - j\}$, where the $i$th source token is aligned to the $j$th target token. 
If $i - j$ is large, it means in order to translate the $j$th target token, the model may need to read more until seeing the $i$th source token. 
Then we calculate the monotonic level of each example as the averaged alignment position shift over the number of aligned tokens, \emph{i.e.}, 
\begin{equation}
    \mathbf{M} = \frac{1}{|\mathbf{A}|}\sum_{(i,j) \in \mathbf{A}} \max\{0, i - j\}.
\end{equation} 
where $\mathbf{M}$ denotes monotonic level and $\mathbf{A}$ represents aligned pairs.
We evenly divide the test set into three groups (Easy, Medium, and Hard) according to different monotonicity levels. 
For each group, we evaluate different inference methods and report the results in Table \ref{tab:reorder}. 
As we explained in \ref{apd:al}, it is almost impossible to guarantee the same AL for different inference methods. 
For a fair comparison, we try our best to set the AL of different methods to be approximately equal. 
We can see our inference strategies show a significant advantage on the non-monotonic examples (Medium and Hard groups).

\section{Conclusion}
In this paper, we examine streaming speech translation from a new perspective. 
We investigate the effects of the input mismatch between offline-training and online-decoding.
We find that the representations at the ending positions in the streaming input are particularly poor, directly impacting the translation quality.
We propose FAST, which introduces future contexts to improve these representations during training and testing via FAD and FAI, respectively.
Experiments and analysis demonstrate their effectiveness in bridging the representation gap between full speech encoding and partial streaming encoding. 
Furthermore, our methods can be generally beneficial to streaming speech translation models that are based on Wav2Vec2.0. 
In the future, we will explore the relevant method independent on Wav2Vec2.0.

\section{Limitations}

Our proposed method is built upon the Wav2Vec2.0 base model, whose superior representation power has been shown to enhance the performance of offline ST models. 
Nevertheless, it should be noted that its parameters are considerably large, approximately 95M. 
This may lead to increased computational costs during training and inference. 
If we want to extend the model to the long context audio (similar to the document level machine translation \cite{zhang2020long}), we have to explore the future work in our conclusion.

The CIF module for detecting the acoustic boundary is optimized from the weakly supervised signal -- total length of text tokens. 
In streaming inference, the boundary detector is not guaranteed to predict accurate boundaries. 
In other words, it is not guaranteed to align each text token with detected boundaries during online inference. 
However, due to the good performance of overall translation quality, we hypothesize that these boundaries may represent some meaningful acoustic (or phrase-like) units. 
The underlying meaning should be another future work to explore.

\section*{Ethics Statement}
After careful review, to the best of our knowledge, we have not violated the \href{https://www.aclweb.org/portal/content/acl-code-ethics}{ACL Ethics Policy}.
Our experiments are based on the open-sourced dataset that is widely used in academia, and there is no violation for this dataset.
Our writing is completely based on the authors without plagiarism.

\section*{Acknowledgements}
We would like to thank all the anonymous reviewers for the insightful and helpful comments.
This work was supported by University-Industry Cooperation Programs of Fujian Province of China (No. 2023H6001), Major Scientific Research Project of the State Language Commission in the 13th Five-Year Plan (Grant no. WT135-38), National Natural Science Foundation of China (No. 62076211), and Alibaba Group through Alibaba Research Intern Program.

\bibliography{anthology,custom}

\newpage
\appendix

\begin{algorithm*}[t]
\caption{Pseudocode of FAI strategy strategy in a PyTorch-like style.}
\label{algo:mask}
\definecolor{codeblue}{rgb}{0.25,0.5,0.5}
\lstset{
  backgroundcolor=\color{white},
  basicstyle=\fontsize{7.2pt}{7.2pt}\ttfamily\selectfont,
  columns=fullflexible,
  breaklines=true,
  captionpos=b,
  commentstyle=\fontsize{7.2pt}{7.2pt}\color{codeblue},
  keywordstyle=\fontsize{7.2pt}{7.2pt},
  escapeinside=``,
}
\begin{lstlisting}[language=python]
# model: an offline-trained ST model consists of a acoustic encoder Wav2vec2.0, a token boundary detector, a semantic encoder, and a decoder
# m: mask length, K: wait lagging, audio: audio waveform
# mask_emb: pre-trained mask embedding in Wav2vec 

N = 0   # the number of source text tokens
x = []  # streaming audio prefix
y = []  # translations
mask_embs = mask_emb.repate(m, 1)  # mask embeddings: `\fontsize{6.8pt}{6.8pt}\color{codeblue}{$m \times d$}`
while y[-1] != "<eos>":
    if x == audio:  # audio has been read
        y = y + model(a,y)  # write new target token
    elif N - len(y) < K:  # wait K detected source tokens
        x = x + read(audio)  # incrementally read audio 
        c = model.wav2vec2.cnn(x)  # audio tokens `\fontsize{6.8pt}{6.8pt}\color{codeblue}{$\tau\times d$}`
        
        c = torch.cat((c, mask_embs), dim=0) # concatenate audio tokens and mask embeddings, `\fontsize{6.8pt}{6.8pt}\color{codeblue}{$(\tau+m)\times d$}`
        a = model.wav2vec2.encoder(c)  # audio representations, `\fontsize{6.8pt}{6.8pt}\color{codeblue}{$(\tau+m)\times d$}`
        a = a[:a.shape[0] - m,:]  # discard the predicted representations, `\fontsize{6.8pt}{6.8pt}\color{codeblue}{$\tau \times d$}`
        
        if model.token_detector(a):  # source text token boundary is detected
            N += 1
    else:
        h = model.semantic_encoder(a)
        y = y + model.decoder(h, y)  # write new target token
\end{lstlisting}
\end{algorithm*}

\section{Data Statistics}
\label{apd:data_statistics}
We evaluate our model on MuST-C V1 English-German (EnDe), English-Spanish (EnEs) and English-French (EnFr) datasets \citep{di-gangi-etal-2019-must}.
For training set, we follow \citet{dong-etal-2022-learning} to filter out short speech of less than 1000 frames (62.5ms) and long speech of more than 480,000 frames (30s).
The statistics of different language pairs are illustrated in Table \ref{tab:data_statistics}.
\begin{table}[h]
\centering
\small
\setlength\tabcolsep{8pt}
\renewcommand\arraystretch{1.2}
\begin{tabular}{lccc}
\toprule
\textbf{split} & \textbf{EnDe} & \textbf{EnEs} & \textbf{EnFr} \\ \midrule
train & 225,271 & 260,041 & 269,248 \\
dev & 1,418 & 1,312 & 1,408 \\
tst-COMMON & 2,641 & 2,502 & 2,632 \\ \bottomrule
\end{tabular}
\caption{Number of samples for each split of MuST-C datasets.}
\label{tab:data_statistics}
\end{table}

\section{Additional Preliminary Analysis}
\label{apd:additional_analysis}

\subsection{Which part of streaming speech representation is worse?}
\label{sec:which_part}
To further verify that only the representation of the end position in streaming speech is poor,
we calculate the cosine similarity $s_{t,t^{\prime}}$ between the speech representation at the $t^{\prime}$-th ($t^{\prime} \leq t$) position in the $t$-th streaming audio input $\mathbf{\hat{x}}_t$ and the speech representation at the same position in the full encoding. 
Then we average the cosine similarities over the sentences in dataset $\mathcal{B}$ to obtain robust statistics. 
\begin{equation}
\begin{aligned}
\text{For $t^\prime \leq t$}, \ \Bar{s}_{t,t^{\prime}} &= \frac{1}{|\mathcal{B}_t|}\sum\limits_{\mathbf{x}\in\mathcal{B}_t} s_{t,t^\prime} (\mathbf{x}) \\
&= \frac{1}{|\mathcal{B}_t|}\sum\limits_{\mathbf{x}\in\mathcal{B}_t} \operatorname{cos}(\hat{a}_{t,t^{\prime}}, a_{t^{\prime}}), 
\label{eq:audio_cos1}
\end{aligned}
\end{equation} 
where $\mathcal{B}_t = \{\mathbf{x}: | \mathbf{x} | \geq t\}$ contains the audio inputs with length no shorter than $t$.

\begin{figure*}[t]
\centering
\subfigure[$\Bar{s}_{t,t^{\prime}}$ of the first three positions]{
\includegraphics[width=0.32\linewidth]{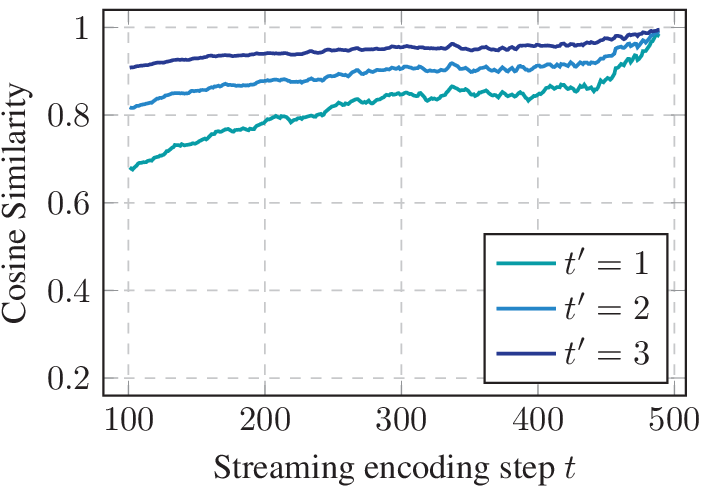}
}
~
\subfigure[$\Bar{s}_{t,t^{\prime}}$ of middle three positions]{
\includegraphics[width=0.3\linewidth]{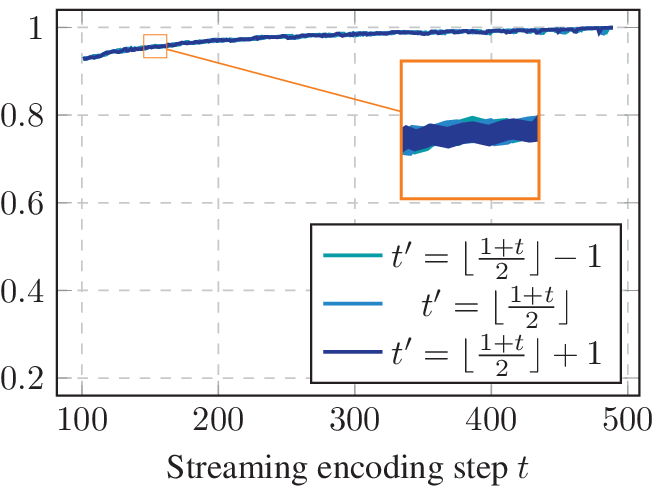}
}
~
\subfigure[$\Bar{s}_{t,t^{\prime}}$ of the last three positions]{
\includegraphics[width=0.3\linewidth]{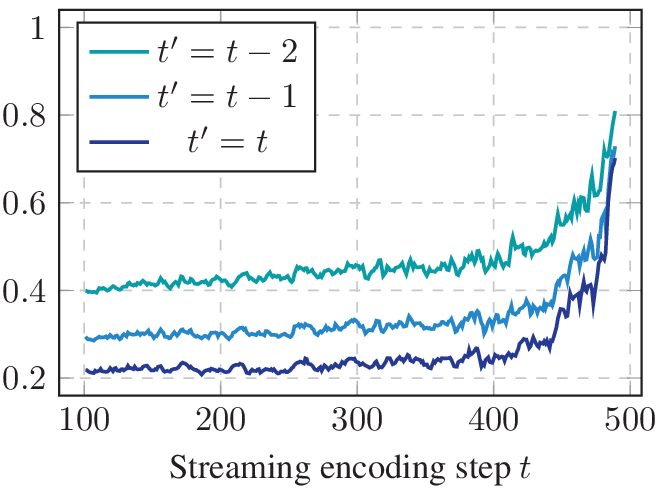}
}
\caption{The average cosine similarity $\Bar{s}_{t,t^{\prime}}$ of the first three ($t^{\prime}=1,2,3$), middle three ($t^{\prime}=\lfloor \frac{1+t}{2} \rfloor-1, \lfloor \frac{1+t}{2} \rfloor, \lfloor \frac{1+t}{2} \rfloor + 1$), and last three ($t^{\prime}=t-2,t-1,t$) positions for each encoding step $t$.}
\label{fig:part}
\end{figure*}

We empirically compare the averaged cosine similarity at the beginning, middle, and end positions of the speech representations. 
Figure \ref{fig:part} shows $\Bar{s}_{t,t^{\prime}}$ of the first three ($t^{\prime}=1,2,3$), middle three ($t^{\prime}=\lfloor \frac{1+t}{2} \rfloor-1, \lfloor \frac{1+t}{2} \rfloor, \lfloor \frac{1+t}{2} \rfloor + 1$), and last three ($t^{\prime}=t-2, t-1, t$) positions for each encoding step $t$. 
At the beginning and middle positions, the averaged cosine similarity $\Bar{s}_{t,t^{\prime}}$ is greater than 0.8 except $t^\prime=1$, indicating that the representations at such positions in the partial streaming input are close to those in the full speech. 
Note that $t^\prime=1$ with a slightly lower similarity won't hurt the performance much, because in practice it is almost impossible to apply \textit{wait}-1 policy (only read 20\emph{ms} speech input) in streaming ST. 
However, the $\Bar{s}_{t,t^{\prime}}$ declines significantly for the end positions, especially for the last one. 
In addition, we observe that as $t$ becomes larger, the streaming input will gradually approximate the full speech input, then the gap of the speech representation between the offline and the online input becomes smaller. 
We conclude that \textbf{the representations of the end position in the streaming speech are particularly inferior.}

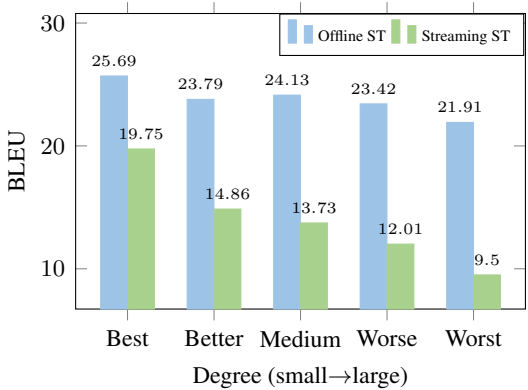
\begin{figure}[t]
\centering
\pgfplotsset{
    width=7.5cm,height=5.5cm,
    every axis y label/.append style={at={(-0.09,0.5)}},
    every axis/.append style={line width=0.4pt},
}
\begin{tikzpicture}[baseline]
\begin{axis}[
    legend style={at={(0.725,1)},
    anchor=north,
    legend columns=2,
    font=\tiny},
    enlargelimits=0.15, 
    font=\small,
    ylabel={BLEU},
    xlabel={Degree (small$\rightarrow$large)},
    xtick=data,
    symbolic x coords={Best,Better,Medium,Worse,Worst},
    xticklabel style={name=tick no\ticknum},
    bar width=10pt,
    nodes near coords,
    nodes near coords align={vertical},
    nodes near coords style={font=\tiny},
    ymax=28,
    ybar=0.1pt,
]

\addplot[ybar, draw=blue1, fill=blue1]
    coordinates {(Best,25.69)(Better,23.79)(Medium,24.13)(Worse,23.42)(Worst,21.91)};

\addplot[ybar, draw=green2, fill=green2]
    coordinates {(Best,19.75)(Better,14.86)(Medium,13.73)(Worse,12.01)(Worst,9.50)};
\legend{Offline ST,Streaming ST}

\end{axis}
\end{tikzpicture}
\caption{Performance with degree of deterioration of the representation at the last position of the streaming speech.}
\label{fig:degree}
\end{figure}

\subsection{Does the poor representation at the last positions of streaming speech affect streaming ST performance?}
\label{apd:degree}

To answer this question, we only calculate the average cosine similarity in the last position for each sample. 
\begin{equation}
\forall \mathbf{x}, \quad \Bar{s}_{-1} (\mathbf{x})= \frac{1}{T}\sum\limits_{t=1}^{t=T} \operatorname{cos}(\hat{a}_{t,t}, a_{t}),
\label{eq:audio_cos3}
\end{equation}
$\Bar{s}_{-1}(\mathbf{x})$ reflects the degree of deterioration of the representation at the last position of the streaming speech. 
We sort the dataset by the value of the degree and divide them evenly into 5 groups to ensure enough samples in each group. The translation quality of each group is shown in Figure \ref{fig:degree}. 
The performance of streaming ST drops close to 10 points as the representation at the last position of the streaming speech becomes worse, while the full-sentence ST fluctuates less than 4 points. 
In addition, the performance gap between the streaming ST and the full-sentence ST becomes larger as the representation at the last position gets worse. In the worse group, the streaming ST is 12.41 points lower than the full-sentence ST. 
Therefore, we conclude that \textbf{the poor representation at the end position of the streaming speech has a strong effect on the translation quality.}

\section{Details of Offline Training} 
\label{apd:details_of_training} 

We use an Adam optimizer with learning rate $1e^{-4}$ and warmup step $10k$.
We decay the learning rate with inverse square root schedule. 

The offline ST model is first trained by a multi-task learning, including ASR and ST tasks. 
A language identity tag is prepended to the target sentence for indicating which task is learned. 
In this stage, the CIF module which is used to detect the acoustic boundary is deactivated, in other words, the CIF module is not trained. 
The main purpose is to learn a better decoder, i.e., a well-trained language model. 
Then, we activate the CIF module such that its parameters are trainable, and continue to train for another several epochs. 
In this stage, only the ST task is learned.

\section{Additional Experiments}

\subsection{Why we use AL rather than \textit{k}?}
\label{apd:al}

In our presented results, we plot the BLEU \emph{v.s.} AL rather than $k$. 
We argue that $k$ is not a fair metric to evaluate the latency. 
In text streaming translation, different tokenization (\emph{e.g.}, different number of BPE operations) will lead to different token boundaries for the same sentence. 
It indicates the $k$ tokens do not necessarily represent the same partial sentence for different BPE methods. 
This situation becomes even severer for speech streaming translation. 
As we have a source text token boundary detector in our model, the first $k$ detected text tokens will represent different lengths of audio frames for different input audios. 
To be precise, the wait-$k$ policy used in our streaming speech translation is actually wait-$k$ detected tokens policy. 
Therefore, we prefer to use AL rather than $k$ as the latency metric in our experiments.

\subsection{How important of the Wav2Vec2.0?}
\label{apd:w2v2}

\begin{figure*}[t]
\pgfplotsset{width=8cm,height=5.5cm,
    every axis y label/.append style={at={(-0.1,0.5)}},
    every axis/.append style={line width=0.6pt},
}
\centering
\subfigure[MuST-C]{
\begin{tikzpicture}[baseline]
\begin{axis}[
    ylabel=Cosine Similarity,
    xlabel=Distance from consumed speech,
    enlargelimits=0.05,
    legend pos=north east,
    legend style={font=\small},
    ymajorgrids=true,
    xmajorgrids=true,
    grid style=dashed,
    font=\small,
    xtick={-10,0,10,20,30,40},
]
\addplot[color=red,mark=o, mark size=0.8pt,line width=0.6pt] coordinates{(-10,0.852)(-9,0.840)(-8,0.827)(-7,0.812)(-6,0.795)(-5,0.778)(-4,0.751)(-3,0.716)(-2,0.673)(-1,0.606)(0,0.568)(1,0.509)(2,0.462)(3,0.422)(4,0.382)(5,0.347)(6,0.319)(7,0.294)(8,0.272)(9,0.254)(10,0.239)(11,0.227)(12,0.217)(13,0.208)(14,0.201)(15,0.195)(16,0.190)(17,0.187)(18,0.183)(19,0.180)(20,0.176)(21,0.173)(22,0.170)(23,0.166)(24,0.162)(25,0.158)(26,0.154)(27,0.150)(28,0.146)(29,0.141)(30,0.137)(31,0.132)(32,0.128)(33,0.124)(34,0.123)(35,0.123)(36,0.123)(37,0.123)(38,0.122)(39,0.119)(40,0.116)(41,0.111)(42,0.106)(43,0.102)(44,0.100)(45,0.117)(46,0.125)(47,0.126)(48,0.127)(49,0.126)};
\addplot[color=poscolor7,mark=o, mark size=0.8pt] coordinates{(-10,0.681)(-9,0.680)(-8,0.676)(-7,0.672)(-6,0.668)(-5,0.665)(-4,0.661)(-3,0.648)(-2,0.596)(-1,0.466)(0,0.354)(1,0.269)(2,0.220)(3,0.183)(4,0.150)(5,0.122)(6,0.099)(7,0.084)(8,0.074)(9,0.068)(10,0.067)(11,0.068)(12,0.069)(13,0.071)(14,0.072)(15,0.072)(16,0.070)(17,0.068)(18,0.065)(19,0.062)(20,0.060)(21,0.060)(22,0.060)(23,0.059)(24,0.058)(25,0.051)(26,0.050)(27,0.049)(28,0.049)(29,0.049)(30,0.050)(31,0.050)(32,0.050)(33,0.048)(34,0.045)(35,0.041)(36,0.037)(37,0.032)(38,0.027)(39,0.025)(40,0.024)(41,0.024)(42,0.026)(43,0.028)(44,0.031)(45,0.031)(46,0.030)(47,0.028)(48,0.022)(49,0.014)}; 
\legend{finetuned-w2v2,pretrained-w2v2}
\end{axis}
\end{tikzpicture}}
~~~
\subfigure[LibriSpeech]{
\begin{tikzpicture}[baseline]
\begin{axis}[
    ylabel=Cosine Similarity,
    xlabel=Distance from consumed speech,
    enlargelimits=0.05,
    legend pos=north east,
    legend style={font=\small},
    ymajorgrids=true,
    xmajorgrids=true,
    grid style=dashed,
    font=\small,
    xtick={-10,0,10,20,30,40},
]
\addplot[color=red,mark=o, mark size=0.8pt] coordinates{(-10,0.8470)(-9,0.836)(-8,0.823)(-7,0.809)(-6,0.793)(-5,0.775)(-4,0.751)(-3,0.719)(-2,0.677)(-1,0.61)(0,0.572)(1,0.516)(2,0.474)(3,0.436)(4,0.399)(5,0.366)(6,0.339)(7,0.315)(8,0.293)(9,0.274)(10,0.258)(11,0.245)(12,0.233)(13,0.222)(14,0.213)(15,0.206)(16,0.200)(17,0.194)(18,0.190)(19,0.185)(20,0.181)(21,0.177)(22,0.173)(23,0.169)(24,0.166)(25,0.162)(26,0.158)(27,0.155)(28,0.151)(29,0.148)(30,0.144)(31,0.141)(32,0.139)(33,0.136)(34,0.136)(35,0.136)(36,0.136)(37,0.135)(38,0.134)(39,0.131)(40,0.130)(41,0.128)(42,0.126)(43,0.124)(44,0.121)(45,0.117)(46,0.115)(47,0.114)(48,0.116)(49,0.114)}; 
\addplot[color=poscolor7,mark=o, mark size=0.8pt] coordinates{(-10,0.639)(-9,0.635)(-8,0.631)(-7,0.626)(-6,0.620)(-5,0.616)(-4,0.606)(-3,0.582)(-2,0.495)(-1,0.351)(0,0.236)(1,0.170)(2,0.138)(3,0.113)(4,0.092)(5,0.072)(6,0.055)(7,0.042)(8,0.034)(9,0.028)(10,0.025)(11,0.025)(12,0.026)(13,0.028)(14,0.030)(15,0.032)(16,0.032)(17,0.032)(18,0.031)(19,0.031)(20,0.030)(21,0.030)(22,0.030)(23,0.030)(24,0.029)(25,0.025)(26,0.024)(27,0.024)(28,0.024)(29,0.024)(30,0.024)(31,0.023)(32,0.022)(33,0.018)(34,0.012)(35,0.005)(36,-0.003)(37,-0.010)(38,-0.013)(39,-0.014)(40,-0.013)(41,-0.010)(42,-0.005)(43,-0.001)(44,0.001)(45,-0.001)(46,-0.004)(47,-0.014)(48,-0.023)(49,-0.001)}; 
\legend{finetuned-w2v2,pretrained-w2v2}
\end{axis}
\end{tikzpicture}}
\caption{We measure the accuracy of predicted context by calculating the cosine similarity between every predicted future representation and full speech representations at the same position. }
\label{fig:mask_acc}
\end{figure*}
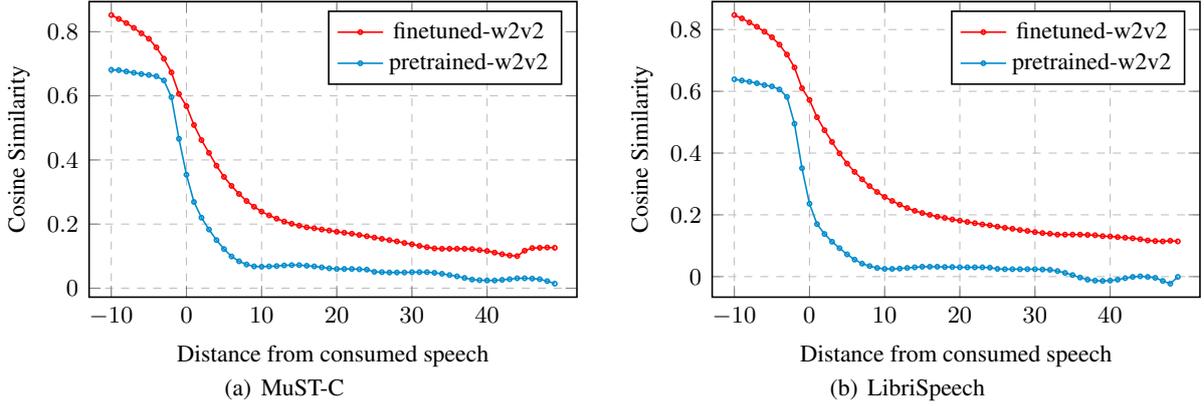

As we mentioned in the main text, the special audio token ``mask" in Wav2Vec2.0 is pre-trained on the Librispeech dataset to reconstruct the corresponding feature conditional on unmasked context via the contrastive task. 
In our experiments, we didn't include contrastive learning as the auxiliary task in the downstream ST training. 
And in our FAI inference, we directly leverage the mask embeddings as the future context by appending them to the streaming input. 
However, we found the speech representations after ST training becomes even better. 
Particularly, we calculate the cosine similarity between every predicted future representation and full speech representations at the same position, and the results are illustrated in Figure \ref{fig:mask_acc}. 
On either the Librispeech or the MuST-C audio test set, the fine-tuned Wav2Vec2.0 can produce better speech representations from the masking inputs.

\subsection{Why $m>10$?}
Based on the analysis in Section 3, we observed that the representations of the last 10 positions of the streaming speech are poorer. For example, the speech representations $\mathbf{\hat{a}}_{t-10:t}$ for streaming speech $\mathbf{c}_{1:t}$ of length $t$ are poor. Similarly, in FKD for a teacher's streaming speech input $\mathbf{c}_{1:t+m}$ of length $t+m$, the speech representations $\mathbf{\hat{a}}_{t+m-10:t+m}$ are always suboptimal. Hence, not all $t+m$ speech representations can be utilized as teachers, only the first $t$ speech representations are taken into account for loss calculation. If $m < 10$, $t+m-10$ will be smaller than $t$, and the representations $\mathbf{\hat{a}}_{t-10+m:t}$ will also be of inferior quality, making the representation $\mathbf{\hat{a}}_{t-10+m:t}$ a poor teacher. Thus, $m$ needs to be greater than 10 for high quality teachers.

\subsection{Why are all predicted features discarded?}
\label{apd:discard}
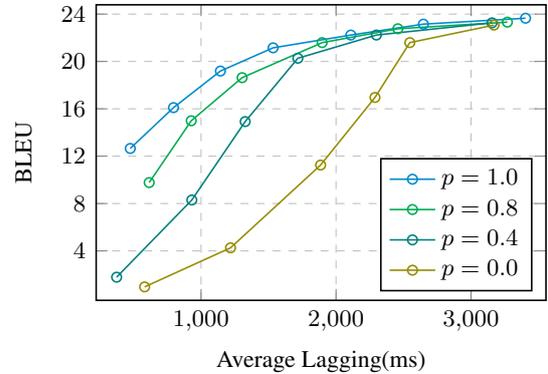
\begin{figure}[t]

\pgfplotsset{width=7.5cm,height=5.5cm,
    every axis y label/.append style={at={(-0.12,0.5)}},
    every axis/.append style={line width=0.6pt},
}
\centering
\begin{tikzpicture}[baseline]
\begin{axis}[
    ylabel=BLEU,
    xlabel=Average Lagging(ms),
    enlargelimits=0.05,
    legend pos=south east,
    legend columns=1,
    legend style={font=\small},
    font=\small,
    xmajorgrids=true,
    ymajorgrids=true,
    grid style=dashed,
    ytick={4,8,12,16,20,24},
]
\addplot[color=poscolor7,mark=o, line width=0.6pt, mark size=1.8pt] coordinates {(475.48,12.65)(795.84,16.1)(1143.37,19.19)(1533.63,21.15)(2109.41,22.23)(2647.03,23.15)(3403.9,23.65)}; 
\addplot[color=poscolor8,mark=o, line width=0.6pt, mark size=1.8pt] coordinates {(615.63,9.77)(927.32,14.99)(1304.13,18.63)(1897.08,21.59)(2457.97,22.76)(3269.3,23.32)};
\addplot[color=teal,mark=o, line width=0.6pt, mark size=1.8pt] coordinates {(374.32,1.78)(931.56,8.3)(1326.92,14.92)(1717.09,20.27)(2298.95,22.23)(3155.64,23.26)};
\addplot[color=olive,mark=o, line width=0.6pt, mark size=1.8pt] coordinates {(581.74,0.96)(1218.34,4.26)(1885.57,11.25)(2288.62,16.96)(2546.59,21.59)(3170.2,23.07)};
\legend{$p=1.0$,$p=0.8$,$p=0.4$,$p=0.0$}
\end{axis}
\end{tikzpicture}
\caption{BLEU \emph{v.s.} AL on different $p$.}
\label{fig:discard}
\end{figure}
In FAI strategy, all the output representations corresponding to the $m=50$ masking tokens will be discarded, because we have demonstrated that the representations at the ending positions are inferior. 
However, as shown in \ref{fig:mask_acc}, the first 10 predicted representations are not as bad as the next 40. 
Therefore, on the EnDE test set, we also conduct another streaming ST inference by appending different numbers of predicted context to the original speech representations. 
We use discard rate $p$ to measure the number of appending features. 
When $p=1.0$, all predicted features are discarded and it reduces to the standard FAI inference. 
In Figure \ref{fig:discard}, we compare the streaming speech translation quality between regular FAI and its variant. 
It is concluded that the predicted future context is too noisy and harmful to the performance.

\subsection{Additional Results on EnDe/Es and EnFr}
\label{apd:appendix_more_results}

\begin{figure*}[!tbp]
\pgfplotsset{width=7.5cm,height=5.8cm,
    every axis y label/.append style={at={(-0.1,0.5)}},
    every axis/.append style={line width=0.6pt},
}
\centering
\subfigure[EnDe]{
\begin{tikzpicture}[baseline]
\begin{axis}[
    ylabel=BLEU,
    xlabel=Average Proportion,
    enlargelimits=0.04,
    font=\small,
    legend pos=south east,
    legend style={font=\scriptsize},
    xmajorgrids=true,
    ymajorgrids=true,
    grid style=dashed,
    ytick={0,4,8,12,16,20,24},
]
\addplot[color=black,dashed,line width=0.8pt] coordinates {(0.1,23.53)(1,23.53)};
\addplot[color=maincolor,mark=*, mark size=1.8pt] coordinates {(0.13,0.02)(0.32,1.68)(0.51,7.79)(0.65,13.31)(0.78,18.72)(0.85,20.67)(0.92,22.33)(0.97,23.16)(1.0,23.29)}; 
\addplot[color=maincolor2,mark=*, mark size=1.8pt,line width=0.8pt] coordinates {(0.3,5.94)(0.53,12.65)(0.63,16.1)(0.7,19.19)(0.76,21.15)(0.83,22.23)(0.88,23.15)(0.93,23.65)(0.97,23.42)};
\addplot[color=maincolor3,mark=*, mark size=1.8pt,line width=0.8pt] coordinates {(0.54,12.69)(0.61,14.78)(0.67,17.71)(0.73,19.67)(0.78,21.36)(0.83,22.51)(0.88,22.84)(0.92,23.36)(0.97,23.55)};

\legend{offline (greedy),Baseline (\textbf{B}),\textbf{B} + FAI,\textbf{B} + FAD + FAI (\textbf{FAST})}
\end{axis}
\end{tikzpicture}}
~
\subfigure[EnDe]{
\begin{tikzpicture}[baseline]
\begin{axis}[
    xlabel=Differentiable Average Lagging,
    enlargelimits=0.04,
    font=\small,
    legend pos=south east,
    legend style={font=\scriptsize},
    xmajorgrids=true,
    ymajorgrids=true,
    grid style=dashed,
    ytick={0,4,8,12,16,20,24},
]
\addplot[color=black,dashed,line width=0.8pt] coordinates {(360,23.53)(6000,23.53)};
\addplot[color=maincolor,mark=*, mark size=1.8pt] coordinates {(359,0.02)(656,1.68)(1032,7.79)(1531,13.31)(2234,18.72)(2788,20.67)(3559,22.33)(4576,23.16)(5791,23.29)}; 
\addplot[color=maincolor2,mark=*, mark size=1.8pt,line width=0.8pt] coordinates {(494,5.94)(928,12.65)(1223,16.1)(1559,19.19)(1928,21.15)(2476,22.23)(2974,23.15)(3678,23.65)(4625,23.42)}; 
\addplot[color=maincolor3,mark=*, mark size=1.8pt,line width=0.8pt] coordinates {(731,12.69)(1009,14.78)(1327,17.71)(1655,19.67)(1991,21.36)(2483,22.51)(2932,22.84)(3581,23.36)(4510,23.55)}; 

\legend{offline (greedy),Baseline (\textbf{B}),\textbf{B} + FAI,\textbf{B} + FAD + FAI (\textbf{FAST})}
\end{axis}
\end{tikzpicture}}
~
\subfigure[EnEs]{
\begin{tikzpicture}[baseline]
\begin{axis}[
    ylabel=BLEU,
    xlabel=Average Proportion,
    enlargelimits=0.04,
    font=\small,
    legend pos=south east,
    legend style={font=\scriptsize},
    xmajorgrids=true,
    ymajorgrids=true,
    grid style=dashed,
    ytick={0,4,8,12,16,20,24,28},
]
\addplot[color=black,dashed,line width=0.8pt] coordinates {(0.34,27.81)(1,27.81)};
\addplot[color=maincolor,mark=*, mark size=1.8pt] coordinates {(0.34,2.39)(0.4,4.09)(0.55,11.37)(0.7,20.31)(0.79,24.62)(0.87,27.04)(0.91,27.74)(0.96,27.88)(0.99,27.76)}; 
\addplot[color=maincolor2,mark=*, mark size=1.8pt,line width=0.8pt] coordinates {(0.35,8.38)(0.59,17.86)(0.7,22.71)(0.78,25.92)(0.83,27.15)(0.89,27.8)(0.93,28.04)(0.96,27.88)(0.99,27.71)};
\addplot[color=maincolor3,mark=*, mark size=1.8pt,line width=0.8pt] coordinates {(0.58,18.34)(0.65,21.53)(0.73,24.78)(0.79,26.4)(0.84,27.24)(0.89,28.02)(0.92,27.98)(0.96,28.23)(0.98,28.09)};
\legend{offline (greedy),Baseline (\textbf{B}),\textbf{B} + FAI,\textbf{B} + FAD + FAI (\textbf{FAST})}
\end{axis}
\end{tikzpicture}}
~
\subfigure[EnEs]{
\begin{tikzpicture}[baseline]
\begin{axis}[
    xlabel=Differentiable Average Lagging,
    enlargelimits=0.04,
    font=\small,
    legend pos=south east,
    legend style={font=\scriptsize},
    xmajorgrids=true,
    ymajorgrids=true,
    grid style=dashed,
    ytick={4,8,12,16,20,24,28},
]
\addplot[color=black,dashed,line width=0.8pt] coordinates {(640,27.81)(4500,27.81)};
\addplot[color=maincolor,mark=*, mark size=1.8pt] coordinates {(1007,2.39)(1054,4.09)(1239,11.37)(1700,20.31)(2215,24.62)(2947,27.04)(3513,27.74)(4260,27.88)(5157,27.76)}; 
\addplot[color=maincolor2,mark=*, mark size=1.8pt,line width=0.8pt] coordinates {(641,8.38)(1181,17.86)(1589,22.71)(2037,25.92)(2500,27.15)(3114,27.8)(3630,28.04)(4328,27.88)(5181,27.71)}; 
\addplot[color=maincolor3,mark=*, mark size=1.8pt,line width=0.8pt] coordinates {(860,18.34)(1232,21.53)(1629,24.78)(2056,26.4)(2471,27.24)(3040,28.02)(3550,27.98)(4214,28.23)(5082,28.09)}; 
\legend{offline (greedy),Baseline (\textbf{B}),\textbf{B} + FAI,\textbf{B} + FAD + FAI (\textbf{FAST})}
\end{axis}
\end{tikzpicture}}
~
\subfigure[EnFr]{
\label{fig:en-fr-ap}
\begin{tikzpicture}[baseline]
\begin{axis}[
    ylabel=BLEU,
    xlabel=Average Proportion,
    enlargelimits=0.04,
    font=\small,
    legend pos=south east,
    legend style={font=\scriptsize},
    xmajorgrids=true,
    ymajorgrids=true,
    grid style=dashed,
    ytick={6,12,18,24,30},
]
\addplot[color=black,dashed,line width=0.6pt] coordinates {(0.35,33.63)(0.95,33.63)};
\addplot[color=maincolor,mark=*, mark size=1.8pt] coordinates {(0.35,3.27)(0.38,3.62)(0.46,6.95)(0.59,15.1)(0.69,21.66)(0.79,27.4)(0.86,29.89)(0.92,31.7)(0.97,33.09)}; 
\addplot[color=maincolor2,mark=*, mark size=1.8pt,line width=0.6pt] coordinates {(0.44,14.45)(0.54,17.61)(0.61,20.63)(0.69,25.87)(0.76,28.95)(0.83,31.47)(0.88,32.68)(0.93,33.11)(0.97,33.54)};
\addplot[color=maincolor3,mark=*, mark size=1.8pt,line width=0.6pt] coordinates {(0.54,19.15)(0.6,22.31)(0.67,25.78)(0.73,28.7)(0.78,30.45)(0.84,32.35)(0.88,33.03)(0.92,33.77)(0.97,33.99)}; 

\legend{offline (greedy),Baseline (\textbf{B}),\textbf{B} + FAI,\textbf{B} + FAD + FAI (\textbf{FAST})}
\end{axis}
\end{tikzpicture}}
~
\subfigure[EnFr]{
\label{fig:en-fr-dal}
\begin{tikzpicture}[baseline]
\begin{axis}[
    xlabel=Differentiable Average Lagging,
    enlargelimits=0.04,
    font=\small,
    legend pos=south east,
    legend style={font=\scriptsize},
    xmajorgrids=true,
    ymajorgrids=true,
    grid style=dashed,
    ytick={6,12,18,24,30},
]
\addplot[color=black,dashed,line width=0.6pt] coordinates {(630,33.63)(4100,33.63)};
\addplot[color=maincolor,mark=*, mark size=1.8pt] coordinates {(997,3.27)(1016,3.62)(1092,6.95)(1300,15.1)(1630,21.66)(2222,27.4)(2741,29.89)(3462,31.7)(4435,33.09)};
\addplot[color=maincolor2,mark=*, mark size=1.8pt,line width=0.6pt] coordinates {(632,14.45)(852,17.61)(1127,20.63)(1456,25.87)(1810,28.95)(2362,31.47)(2838,32.68)(3515,33.11)(4454,33.54)};
\addplot[color=maincolor3,mark=*, mark size=1.8pt,line width=0.6pt] coordinates {(705,19.15)(985,22.31)(1293,25.78)(1616,28.7)(1943,30.45)(2418,32.35)(2850,33.03)(3473,33.77)(4376,33.99)};

\legend{offline (greedy),Baseline (\textbf{B}),\textbf{B} + FAI,\textbf{B} + FAD + FAI (\textbf{FAST})}
\end{axis}
\end{tikzpicture}}
\caption{The translation quality (BLEU) against the latency metrics (AP, DAL) on the tst-COMMON set of MuST-C EnDe, EnEs and EnFr dataset.}
\label{fig:expand_results}
\end{figure*}
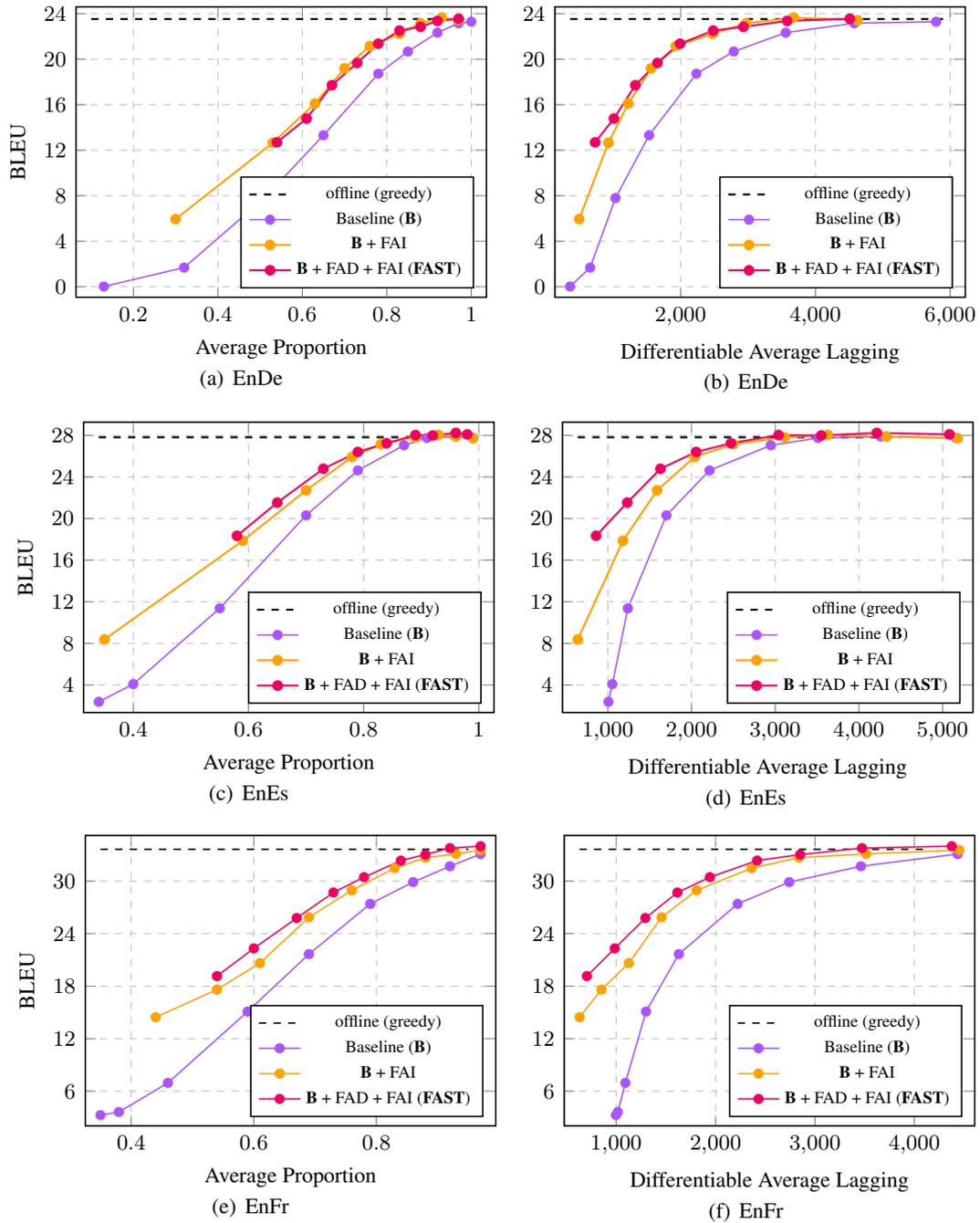

In this section, we evaluate our methods with other latency metrics AP and DAL. 
The AP-BLEU and DAL-BLEU curves on the MuST-C EnDe, EnEs, and EnFr tst-COMMON sets are shown in Figure \ref{fig:expand_results}. 
For three language pairs, our methods can consistently improve the baseline by a large margin. 

\section{Numeric Results for the Figures}
\label{apd:appendix_numeric_results}

\begin{table*}[!ht]
\begin{center}
\renewcommand\arraystretch{1.2}
\resizebox{\textwidth}{!}{
\begin{tabular}{lccccccccccccc}
\toprule
\multirow{2}{*}{Model}    & \multirow{2}{*}{Lagging ($k$)} & \multicolumn{4}{c}{En-De}  & \multicolumn{4}{c}{En-Es}  & \multicolumn{4}{c}{En-Fr}  \\ \cmidrule(lr){3-6} \cmidrule(lr){7-10}\cmidrule(lr){11-14}
                          &                              & AL   & AP   & DAL  & BLEU  & AL   & AP   & DAL  & BLEU  & AL   & AP   & DAL  & BLEU  \\ \cmidrule(lr){1-1} \cmidrule(lr){2-2} \cmidrule(lr){3-6} \cmidrule(lr){7-10}\cmidrule(lr){11-14}
\multirow{9}{*}{Baseline} & 1                            & 178  & 0.13 & 359  & 0.02  & 295  & 0.34 & 1007 & 2.39  & 288  & 0.35 & 997  & 3.27  \\
                          & 3                            & 483  & 0.32 & 656  & 1.68  & 543  & 0.40  & 1054 & 4.09  & 463  & 0.38 & 1016 & 3.62  \\
                          & 5                            & 659  & 0.42 & 821  & 4.34  & 882  & 0.55 & 1239 & 11.37 & 693  & 0.46 & 1092 & 6.95  \\
                          & 7                            & 867  & 0.51 & 1032 & 7.79  & 1361 & 0.70  & 1700 & 20.31 & 1028 & 0.59 & 1300 & 15.10  \\
                          & 9                            & 1295 & 0.65 & 1531 & 13.31 & 1848 & 0.79 & 2215 & 24.62 & 1406 & 0.69 & 1630 & 21.66 \\
                          & 12                           & 1939 & 0.78 & 2234 & 18.72 & 2572 & 0.87 & 2947 & 27.04 & 1972 & 0.79 & 2222 & 27.40  \\
                          & 15                           & 2505 & 0.85 & 2788 & 20.67 & 3171 & 0.91 & 3513 & 27.74 & 2495 & 0.86 & 2741 & 29.89 \\
                          & 20                           & 3312 & 0.92 & 3559 & 22.33 & 3988 & 0.96 & 4260 & 27.88 & 3245 & 0.92 & 3462 & 31.70  \\
                          & 30                           & 4410 & 0.97 & 4576 & 23.16 & 5012 & 0.99 & 5157 & 27.76 & 4283 & 0.97 & 4435 & 33.09 \\ \cmidrule(lr){1-1} \cmidrule(lr){2-2} \cmidrule(lr){3-6} \cmidrule(lr){7-10}\cmidrule(lr){11-14}
\multirow{9}{*}{+ FAI}      & 1                            & 150  & 0.30  & 494  & 5.94  & 347  & 0.35 & 641  & 8.38  & 285  & 0.44 & 632  & 14.45 \\
                          & 3                            & 475  & 0.53 & 928  & 12.65 & 775  & 0.59 & 1181 & 17.86 & 505  & 0.54 & 852  & 17.61 \\
                          & 5                            & 796  & 0.63 & 1223 & 16.10  & 1162 & 0.70  & 1589 & 22.71 & 805  & 0.61 & 1127 & 20.63 \\
                          & 7                            & 1143 & 0.70  & 1559 & 19.19 & 1608 & 0.78 & 2037 & 25.92 & 1154 & 0.69 & 1456 & 25.87 \\
                          & 9                            & 1534 & 0.76 & 1928 & 21.15 & 2076 & 0.83 & 2500 & 27.15 & 1498 & 0.76 & 1810 & 28.95 \\
                          & 12                           & 2109 & 0.83 & 2476 & 22.23 & 2736 & 0.89 & 3114 & 27.80  & 2060 & 0.83 & 2362 & 31.47 \\
                          & 15                           & 2647 & 0.88 & 2974 & 23.15 & 3301 & 0.93 & 3630 & 28.04 & 2559 & 0.88 & 2838 & 32.68 \\
                          & 20                           & 3404 & 0.93 & 3678 & 23.65 & 4072 & 0.96 & 4328 & 27.88 & 3280 & 0.93 & 3515 & 33.11 \\
                          & 30                           & 4457 & 0.97 & 4625 & 23.42 & 5045 & 0.99 & 5181 & 27.71 & 4297 & 0.97 & 4454 & 33.54 \\ \cmidrule(lr){1-1} \cmidrule(lr){2-2} \cmidrule(lr){3-6} \cmidrule(lr){7-10}\cmidrule(lr){11-14}
\multirow{9}{*}{FAST}     & 1                            & 41   & 0.54 & 731  & 12.69 & 270  & 0.58 & 860  & 18.34 & 223  & 0.54 & 705  & 19.15 \\
                          & 3                            & 403  & 0.61 & 1009 & 14.78 & 722  & 0.65 & 1232 & 21.53 & 554  & 0.60  & 985  & 22.31 \\
                          & 5                            & 771  & 0.67 & 1327 & 17.71 & 1152 & 0.73 & 1629 & 24.78 & 895  & 0.67 & 1293 & 25.78 \\
                          & 7                            & 1135 & 0.73 & 1655 & 19.67 & 1594 & 0.79 & 2056 & 26.40  & 1224 & 0.73 & 1616 & 28.70 \\
                          & 9                            & 1503 & 0.78 & 1991 & 21.36 & 2031 & 0.84 & 2471 & 27.24 & 1570 & 0.78 & 1943 & 30.45 \\
                          & 12                           & 2036 & 0.83 & 2483 & 22.51 & 2650 & 0.89 & 3040 & 28.02 & 2079 & 0.84 & 2418 & 32.35 \\
                          & 15                           & 2539 & 0.88 & 2932 & 22.84 & 3194 & 0.92 & 3550 & 27.98 & 2541 & 0.88 & 2850 & 33.03 \\
                          & 20                           & 3260 & 0.92 & 3581 & 23.36 & 3943 & 0.96 & 4214 & 28.23 & 3212 & 0.92 & 3473 & 33.77 \\
                          & 30                           & 4305 & 0.97 & 4510 & 23.55 & 4928 & 0.98 & 5082 & 28.09 & 4199 & 0.97 & 4376 & 33.99 \\ \bottomrule
\end{tabular}}
\end{center}
\caption{Numeric results on MuST-C EnDe, EnEs, and EnFr 
tst-COMMON set (Figure \ref{fig:main_results} and \ref{fig:expand_results}). }
\label{tab:main_exp_numeric}
\end{table*}

\begin{table*}[ht]
\setlength\tabcolsep{10pt}
\begin{center}
\begin{tabular}{ccccccccc}
\toprule
\multirow{2}{*}{Lagging ($k$)} & \multicolumn{2}{c}{\textit{w/o $\mathcal{L}_{KD}^{\text{W2V2}}$}} & \multicolumn{2}{c}{\textit{w/o $\mathcal{L}_{KD}^{\text{CIF}}$}} & \multicolumn{2}{c}{\textit{w/o FAI}} & \multicolumn{2}{c}{\textit{w/o mask embeds}} \\ \cmidrule(lr){2-3} \cmidrule(lr){4-5} \cmidrule(lr){6-7}\cmidrule(lr){8-9}
                               & AL         & BLEU       & AL         & BLEU       & AL         & BLEU       & AL         & BLEU        \\ \cmidrule(lr){1-1} \cmidrule(lr){2-3} \cmidrule(lr){4-5} \cmidrule(lr){6-7}\cmidrule(lr){8-9}
1                              & 139        & 12.00      & 756        & 16.78      & 177        & 2.56       & 115        & 10.13       \\
3                              & 533        & 13.76      & 1220       & 20.01      & 390        & 2.80       & 459        & 11.86       \\
5                              & 911        & 16.24      & 1671       & 21.52      & 605        & 4.49       & 836        & 14.01       \\
7                              & 1288       & 18.17      & 2112       & 22.24      & 888        & 9.01       & 1211       & 15.12       \\
9                              & 1682       & 19.08      & 2527       & 22.41      & 1247       & 13.68      & 1588       & 15.76       \\
12                             & 2231       & 19.78      & 3087       & 22.68      & 1812       & 18.22      & 2138       & 16.44       \\
15                             & 2722       & 20.17      & 3562       & 22.73      & 2338       & 20.41      & 2641       & 16.62       \\
20                             & 3434       & 20.43      & 4201       & 22.84      & 3105       & 22.25      & 3363       & 16.75       \\
30                             & 4443       & 20.35      & 4992       & 22.73      & 4217       & 23.36      & 4393       & 16.63      \\ \bottomrule
\end{tabular}
\end{center}
\caption{Numeric results for ablation study (Figure \ref{fig:ablation}).}
\label{tab:ablation}
\end{table*}

\begin{table*}[ht]
\begin{center}
\begin{tabular}{llllllllllll}
\toprule
\multirow{16}{*}{EnDe} & \multicolumn{11}{l}{\textit{\textbf{MU-ST}}} \\ \cmidrule(lr){2-12}
                       & AL          & 1023  & 1424  & 1953  & 2642  & 3621  & 4453  & 5089  & 5754 \\ 
                       & BLEU        & 17.94 & 20.85 & 22.78 & 24.30  & 24.82 & 24.99 & 25.05 & 25.90 \\ \cmidrule(lr){2-12}
                       & \multicolumn{11}{l}{\textit{\textbf{RealTrans}}}   \\ \cmidrule(lr){2-12}
                       & AL          & 1355  & 1838  & 2290  & 2720  & 3106  \\ 
                       & BLEU        & 16.54 & 18.49 & 19.84 & 20.05 & 20.41 \\ \cmidrule(lr){2-12}
                       & \multicolumn{11}{l}{\textit{\textbf{MoSST}}} \\ \cmidrule(lr){2-12}
                       & AL   & 728   & 862   & 1021  & 1689  & 2088  \\
                        & BLEU & 7.07  & 9.04  & 11.52 & 16.44 & 17.31 \\ \cmidrule(lr){2-12}
                       & \multicolumn{11}{l}{\textit{\textbf{ITST}}} \\ \cmidrule(lr){2-12}
                       & AL   & 1449  & 1589  & 1678  & 1778  & 1919  & 2137  & 2371  \\
                        & BLEU & 17.90  & 18.47 & 19.09 & 19.50  & 20.09 & 20.64 & 21.06 \\ 
                        & AL  & 2618  & 2893  & 3193  & 3501  & 3876  & 4557  & 5206  \\
                        & BLEU & 21.64 & 21.80  & 22.02 & 22.27 & 22.51 & 22.62 & 22.71 \\ \midrule
\multirow{11}{*}{EnEs} & \multicolumn{11}{l}{\textit{\textbf{SimulSpeech}}}       \\ \cmidrule(lr){2-12}
                       & AL          & 694   & 1336  & 2169  & 2724  & 3331   \\
                       & BLEU        & 15.02 & 19.92 & 21.58 & 22.42 & 22.49  \\ \cmidrule(lr){2-12}
                       & \multicolumn{11}{l}{\textit{\textbf{RealTrans}}}   \\ \cmidrule(lr){2-12}
                       & AL          & 1047  & 1554  & 2043  & 2514  & 2920  \\
                       & BLEU        & 18.54 & 22.74 & 24.89 & 25.54 & 25.97 \\ \cmidrule(lr){2-12}
                       & \multicolumn{11}{l}{\textit{\textbf{ITST}}} \\ \cmidrule(lr){2-12}
                       & AL   & 960   & 1153  & 1351  & 1621  & 1964  & 2381  & 2643  & 2980  & 3434  & 3983  \\
                        & BLEU & 17.77 & 18.38 & 18.71 & 19.11 & 19.77 & 20.13 & 20.46 & 20.75 & 20.48 & 20.64 \\ \midrule
\multirow{3}{*}{EnFr} & \multicolumn{11}{l}{\textit{\textbf{MMA-SLM}}}       \\ \cmidrule(lr){2-12}
                       & AL          & 701   & 1197  & 1704  \\
                       & BLEU        & 14.86 & 19.79 & 25.16 \\ \bottomrule
\end{tabular}
\end{center}
\caption{Numeric results for baseline systems (Figure \ref{fig:main_results}). The results of \textit{\textbf{MU-ST}} are obtained from \citep{zhang-etal-2022-learning}. The results of \textit{\textbf{SimulSpeech}} and \textit{\textbf{RealTrans}} are obtained from \citep{zeng-etal-2021-realtrans}. The results of \textit{\textbf{MoSST}} are obtained from \citep{dong-etal-2022-learning}. The results of \textit{\textbf{ITST}} are obtained from \citep{zhang-feng-2022-information}. The results of \textit{\textbf{MMA-SLM}} are obtained from \citep{indurthi-etal-2022-language}.}
\label{tab:baseline_numeric}
\end{table*}

\begin{table*}[ht]
\begin{center}
\renewcommand\arraystretch{1.2}
\resizebox{\textwidth}{!}{
\begin{tabular}{ccccccccccccccc}
\toprule
\multirow{2}{*}{Lagging ($k$)} & \multicolumn{2}{c}{$m=5$} & \multicolumn{2}{c}{$m=10$} & \multicolumn{2}{c}{$m=20$} & \multicolumn{2}{c}{$m=30$} & \multicolumn{2}{c}{$m=50$} & \multicolumn{2}{c}{$m=80$} & \multicolumn{2}{c}{$m=100$} \\ \cmidrule(lr){2-3} \cmidrule(lr){4-5} \cmidrule(lr){6-7}\cmidrule(lr){8-9}\cmidrule(lr){10-11}\cmidrule(lr){12-13}
                               \cmidrule(lr){14-15}
                               & AL         & BLEU       & AL         & BLEU        & AL         & BLEU        & AL         & BLEU        & AL         & BLEU        & AL         & BLEU        & AL          & BLEU        \\ \cmidrule(lr){1-1} \cmidrule(lr){2-3} \cmidrule(lr){4-5} \cmidrule(lr){6-7}\cmidrule(lr){8-9}\cmidrule(lr){10-11}\cmidrule(lr){12-13}
                               \cmidrule(lr){14-15}
1                              & 118        & 0.49       & 64         & 5.67        & 99         & 12.67       & 3          & 12.44       & 41         & 12.69       & 85         & 12.78       & 100         & 13.18       \\
3                              & 298        & 8.48       & 306        & 12.10        & 468        & 15.20        & 349        & 14.50        & 403        & 14.78       & 458        & 15.57       & 479         & 15.87       \\
5                              & 629        & 13.84      & 660        & 16.03       & 858        & 18.24       & 717        & 16.87       & 771        & 17.71       & 835        & 17.87       & 845         & 17.91       \\
7                              & 1003       & 17.38      & 1038       & 18.78       & 1237       & 20.23       & 1083       & 19.32       & 1135       & 19.67       & 1205       & 19.97       & 1225        & 20.07       \\
9                              & 1389       & 19.38      & 1424       & 20.2        & 1627       & 21.56       & 1466       & 21.14       & 1503       & 21.36       & 1562       & 21.61       & 1587        & 21.44       \\
12                             & 1957       & 21.46      & 1978       & 21.62       & 2189       & 22.45       & 2001       & 22.19       & 2036       & 22.51       & 2095       & 22.38       & 2109        & 22.47       \\
15                             & 2479       & 22.17      & 2497       & 22.58       & 2695       & 23.02       & 2507       & 22.75       & 2539       & 22.84       & 2588       & 23.08       & 2599        & 23.07       \\
20                             & 3228       & 22.91      & 3231       & 23.14       & 3425       & 23.29       & 3234       & 23.43       & 3260       & 23.36       & 3302       & 23.55       & 3311        & 23.54   \\ \bottomrule
\end{tabular}}
\end{center}
\caption{Numeric results for different lengths future context (Figure \ref{fig:length_bleu}).}
\label{tab:future_length_numeric}
\end{table*}

We also provide the numeric results for Figures \ref{fig:main_results} and \ref{fig:expand_results} in Tables \ref{tab:main_exp_numeric}, and for Figures \ref{fig:ablation} in Table \ref{tab:ablation}, and for Figures \ref{fig:main_results} in Table \ref{tab:baseline_numeric}, for Figure \ref{fig:length_bleu} in Table \ref{tab:future_length_numeric}.

\end{document}